\documentclass[conference]{IEEEtran}
\IEEEoverridecommandlockouts
\usepackage{cite}
\usepackage{amsmath,amssymb,amsfonts}
\usepackage{algorithmic}
\usepackage{graphicx}
\usepackage{textcomp}
\usepackage{xcolor}
\usepackage{multirow}
\usepackage{tabularx}
\usepackage[algo2e]{algorithm2e}
\usepackage{algorithm,algorithmic}
\usepackage{lipsum}
\usepackage{mathtools}
\usepackage{cuted}
\usepackage{graphicx}
\usepackage{caption}
\usepackage{subcaption}
\usepackage{subcaption}
\usepackage{adjustbox}
\usepackage{array}
\usepackage[T1]{fontenc}

\expandafter\def\expandafter\normalsize\expandafter{%
    \normalsize%
    \setlength\abovedisplayskip{-4pt}%
    \setlength\belowdisplayskip{8pt}%
    \setlength\abovedisplayshortskip{-8pt}%
    \setlength\belowdisplayshortskip{2pt}%
}
\newcolumntype{C}[1]{{\centering\let\newline\\\arraybackslash\hspace{0pt}}m{#1}}
\usepackage{graphicx,xcolor,subcaption,lipsum}
\def\BibTeX{{\rm B\kern-.05em{\sc i\kern-.025em b}\kern-.08em
    T\kern-.1667em\lower.7ex\hbox{E}\kern-.125emX}}
\begin{document}

\title{S-E Pipeline: A Vision Transformer (ViT) based Resilient Classification Pipeline for Medical Imaging Against Adversarial Attacks\\
}

\author{\IEEEauthorblockN{Neha A S}
\IEEEauthorblockA{\textit{Dept. of CSE} \\
\textit{Indian Institute of Technology Palakkad}\\
Kerala, India\\
111914005@smail.iitpkd.ac.in}
\and
\IEEEauthorblockN{Vivek Chaturvedi}
\IEEEauthorblockA{\textit{Dept. of CSE} \\
\textit{Indian Institute of Technology Palakkad}\\
Kerala, India \\
vivek@iitpkd.ac.in}
\and
\IEEEauthorblockN{Muhammad Shafique}
\IEEEauthorblockA{\textit{Division of Engineering} \\
\textit{New York University}\\
Abu Dhabi, UAE\\
muhammad.shafique@nyu.edu}
}

\maketitle

\begin{abstract}
Vision Transformer (ViT) is becoming widely popular in automating accurate disease diagnosis in medical imaging owing to its robust self-attention mechanism. However, ViTs remain vulnerable to adversarial attacks that may thwart the diagnosis process by leading it to intentional misclassification of critical disease. In this paper, we propose a novel image classification pipeline, namely, S-E Pipeline, that performs multiple pre-processing steps that allow ViT to be trained on critical features so as to reduce the impact of input perturbations by adversaries. Our method uses a combination of segmentation and image enhancement techniques such as Contrast Limited Adaptive Histogram Equalization (CLAHE), Unsharp Masking (UM), and High-Frequency Emphasis filtering (HFE) as preprocessing steps to identify critical features that remain intact even after adversarial perturbations. The experimental study demonstrates that our novel pipeline helps in reducing the effect of adversarial attacks by 72.22\% for the ViT-b32 model and 86.58\% for the ViT-l32 model. Furthermore, we have shown an end-to-end deployment of our proposed method on the NVIDIA Jetson Orin Nano board to demonstrate its practical use case in modern hand-held devices that are usually resource-constrained.
\end{abstract}
\vspace{-0.2cm}
\begin{IEEEkeywords}
vision transformers, medical imaging, adversarial attacks, defense mechanisms, image enhancement, segmentation
\end{IEEEkeywords}
\vspace{-0.2cm}
\section{Introduction}

Deep learning models are becoming pervasive in automating the diagnosis process in medical imaging. For instance, models such as Convolutional Neural Networks (CNNs), Recurrent Neural Networks (RNNs), and Long Short Term Memory (LSTM) have shown great potential in diagnosing critical diseases accurately using image outputs from medical imaging techniques such as Radiography, CT scans, and Ultrasound \cite{b3, mortazi2018automatically,sultana2020review}. However, these models are vulnerable to adversarial attacks that introduce some noise or perturbations in the input image which are usually imperceptible to human eyes but are sufficient to fool these models and lead to intentional misclassification. This not only thwarts the diagnosis process but can also be used for financial and privacy frauds \cite{ma2021understanding}.
Recently, Vision Transformers (ViTs) have become popular in computer vision applications as they have a very robust self-attention mechanism that allows capturing global relations by topological relationships calculated
image patches. ViTs are fast becoming a first-class deep learning architecture choice in the medical imaging domain as well \cite{wang2023cross, li2023lvit, almalik2022self}. However, similar to the other traditional models like CNNs, ViTs are also vulnerable to adversarial attacks, though to some degree less.

Adversarial attacks are a potential threat to all machine learning models \cite{ma2021understanding}. Adversarial machine learning is a well-researched area in the case of CNNs \cite{fezza2019perceptual} and is now gaining attention in the case of ViTs \cite{mahmood2021robustness}. CNNs and ViTs are prone to white-box adversarial attacks like FGSM \cite{fgsm} and PGD \cite{pgd}, and the attack success rate is enough to distract an established system \cite{9259112, shao2022adversarial}. Existing defense mechanisms including pre-processing-based defenses \cite{9420266} and modifications in the architecture \cite{chang2023enhancing} are studied in the literature but most of the works focus on models processing natural images. Thereby, there exists a gap in defense mechanisms that can work well in the medical imaging domain.

Moreover, the majority of the defense mechanisms proposed are not designed for resource-constrained computing environments. As it is evident that medical imaging devices are also becoming mobile, it is critical that the proposed solutions should be able to perform effectively in resource-constrained computing platforms like edge devices.

In this paper, we have proposed a novel image classification pipeline, the Segmentation-Enhancement (S-E) Pipeline, which in combination with ViT provides a robust medical diagnosis process and remains resilient to some of the well-known state-of-the-art adversarial attacks, FGSM \cite{fgsm} and PGD \cite{pgd}. Our S-E pipeline performs multiple pre-processing steps such as segmentation and image enhancement techniques including CLAHE, UM and HFE.  Experimental results demonstrate that by using this pipeline, we can reduce the effect of FGSM and PGD attacks by 72.22\% \& 36.25\% respectively for ViT-b32, and 86.58\% \& 80.26\% respectively for ViT-l32 models.
Furthermore, as the implementation of the proposed methodology in hardware ensures that it can operate in practical devices, we have provided an end-to-end deployment of our S-E pipeline on the NVIDIA Jetson Orin Nano board.

\textbf{The main contributions of our work are:}
\begin{itemize}
    \item Leveraging segmentation and image enhancement techniques as a defensive pre-processing pipeline against adversarial attacks for CNN and ViT processing chest X-ray images.
  
    \item The best method after experimenting with different parameter values in the pre-processing pipeline gives a maximum of 86.58\% and 84.70\% reduction in the effect of attacks in ViT and CNNs respectively.
      \item Implementation of the proposed method in NVIDIA Jetson Orin Nano board demonstrating its feasibility for deployment in resource-constrained applications.
\end{itemize}

\vspace{-0.2cm}
\section{Related Work}
\vspace{-0.2cm}
Adversarial machine learning for medical images is well-studied in the case of CNNs and its vulnerability to various types of attacks is showcased in various works \cite{finlayson2018adversarial, paschali2018generalizability}. To counter these attacks, several defense mechanisms are also developed \cite{kaviani2022adversarial}. One of the major categories of defense is pre-processing-based defenses where some changes are applied to the adversarial image before forwarding to the model. JPEG compression \cite{dziugaite2016study} is applied to images as a pre-processing step to remove high-frequency components as they may amplify the effect of perturbations. Repositioning pixels in an image using re-scaling and cropping \cite{graese2016assessing}, and local smoothing \cite{Xu_2018} where each pixel is smoothed out using pixel neighborhood are some successful defenses. Non-local smoothing \cite{Xu_2018} smoothes out the current patch using a search window and it may remove perturbations with zero mean noise. Total variation minimization \cite{guo2017countering} is another technique where it minimizes an image’s total variation whereas the flat regions are denoised. Robustness of vision transformers are studied in \cite{aldahdooh2021reveal} where all the above defense mechanisms are tested in ViTs and a general observation is that ViTs are robust compared to CNNs.
\begin{figure}
\centering
    \includegraphics[scale=0.42]{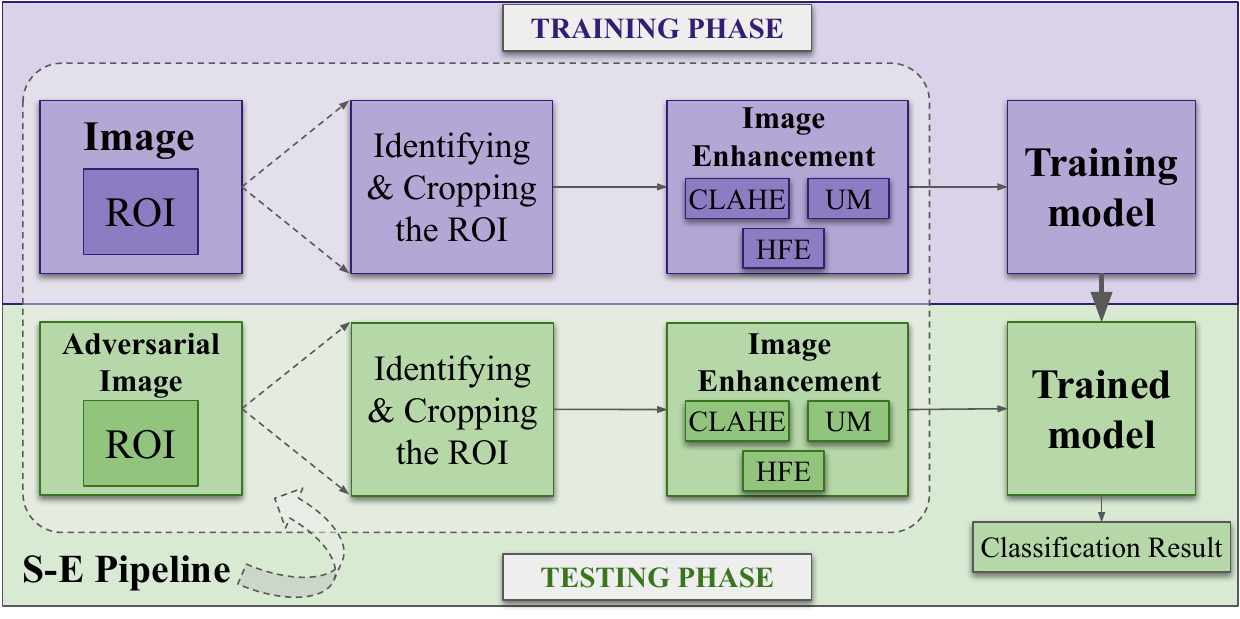}
    \caption{Overall workflow of the proposed system.}
    \label{1}
\end{figure}
In another method for defense, adversarial training, the attacked images are used for re-training the model and it proves to be successful in reducing the effect of attacks. Experimental results of \cite{mo2022adversarial} show that randomly masking perturbations or gradients from attention blocks can improve the robustness of ViTs against adversarial attacks. Another study \cite{electronics11203370} utilizes adversarial training along with the introduction of a module to extract features and to select the size of the perpetual field adaptively. Adversarial training techniques can work efficiently when the attacks used to train the models are used and are often inefficient against new attacks. 

In the medical imaging domain, \cite{chang2023enhancing} introduces the ResNet-SE module using which the model can extract features by focusing on important information from the feature map. A reverse diffusion technique using defensive diffusion \cite{imam2023enhancing} can eliminate adversarial noise from the original image. MedViT \cite{MANZARI2023106791} is a robust ViT developed for disease diagnosis. It is a hybrid model incorporating both local and global features and was tested on the MedMNIST-2D dataset for its performance.

\vspace{-0.1cm}
\section{Proposed Method: S-E Pipeline}

We propose a ViT-based defense mechanism using segmentation and image enhancement techniques, a Segmentation-Enhancement (S-E) pipeline,  for models processing lung X-ray images. The overall working of the proposed method is depicted in Figure \ref{1}. In the training phase, the Region of Interest (ROI) is cropped from the image using segmentation. Then, the image is passed to an image enhancement phase, where any one of the pre-processing techniques, CLAHE, UM, or HFE is used. Then, the image is fed into a ViT classifier for training. The trained model is then saved for the inference part. In the testing phase, the same process is repeated for the adversarial image, except for the usage of the trained model. The adversarial image is segmented to obtain the ROI, after which any one of the aforementioned, CLAHE, UM, or HFE is applied to the segmented image and is finally fed to the trained model for predictions. The workflow of the proposed system is detailed in Algorithm \ref{alg}. Details and workings of each block are explained in the subsections below.

\SetKwInput{KwInput}{Input}                
\SetKwInput{KwOutput}{Output}

\begin{algorithm}[!h]
\DontPrintSemicolon
  \KwInput{X-ray image dataset}
  \KwOutput{Rate of reduction, Improved model}
  \SetKwFunction{FMain}{Main}
   \SetKwFunction{seg}{segment}
   \SetKwFunction{enh}{imgEnhance}
   \SetKwProg{Fn}{Def}{:}{}
   
  \Fn{\seg}{
   seg\_img = U-Net(image) \& image }
  \KwRet seg\_img;

  \Fn{\enh}{
   \If{option = CLAHE}
   {
   Split image to tiles of size 4*4 \\
    Calculate the histogram, contrast limit size=8\\
   enh\_img = Contrast limited image
   }
   \If{option = UM}{
   blur\_img = Apply blur to the image \\
   /* blur: Gaussian, Median, Maximum, Minimum */ \\
   unsharped\_mask = image - blur\_img \\
   enh\_img = image + amount $\times$ (unsharped\_mask) 
   }
   \If{option = HFE}
   {
   enh\_img = (image + (Gaussian Highpass Filter)) $\times$ (Histogram Equalization)
   }}
  \KwRet enh\_img;

  \Fn{\FMain}{
    train\_set, test\_set = 80:20 splitting of dataset \\
    \For{image in train\_set}
    {
    seg\_img = segment(image) \\
    enh\_img = imgEnhance(image, option) \\/* option = CLAHE, UM, HFE */\\
    model = train(model, option) \\/* option = CNN, ViT */\\
    }
    \For{image in $test\_set$}
    {
    $P_Atotal$, $P_Btotal$, $P_{Aadv}total$, $P_{Badv}total$ = 0 \\
    $P_A, P_B$ = model(image) \\ /* Probabilities belonging to class A and B */ \\ 
    adv\_img = adversarial attack(image)\\
    $P_{Aadv}, P_{Badv}$ = model(adv\_image) \\
    $P_Atotal$ = $P_Atotal$ +  $P_A$ \\
    $P_Btotal$ = $P_Btotal$ +  $P_B$ \\
    $P_{Aadv}total$ = $P_{Aadv}total$ + $P_{Aadv}$ \\
    $P_{Badv}total$ = $P_{Badv}total$ + $P_{Badv}$ \\
   }
    $P\_total$ = ( $P_Atotal$ + $P_Btotal$ ) /2 \\
    $P_{adv}\_total$ = ( $P_{Aadv}total$ + $P_{Badv}total$ ) /2 \\
    Diff\_normal = $|P\_total - P_{adv}\_total|$ \\
    
    Similarly, calculate Diff\_pre for S-E pipeline attached model \\
    Rate of Reduction = (Diff\_normal-Diff\_pre)/Diff\_normal 
  }
  
\caption{\textbf{S-E Pipeline workflow} - Segmentation, Enhancement, Model Training, Result Calculations}
\label{alg}
\end{algorithm}

\vspace{-0.1cm}
\subsection{Segmentation}
\vspace{-0.5cm}
\begin{figure}[h]
    \centering
    \includegraphics[scale=0.3]{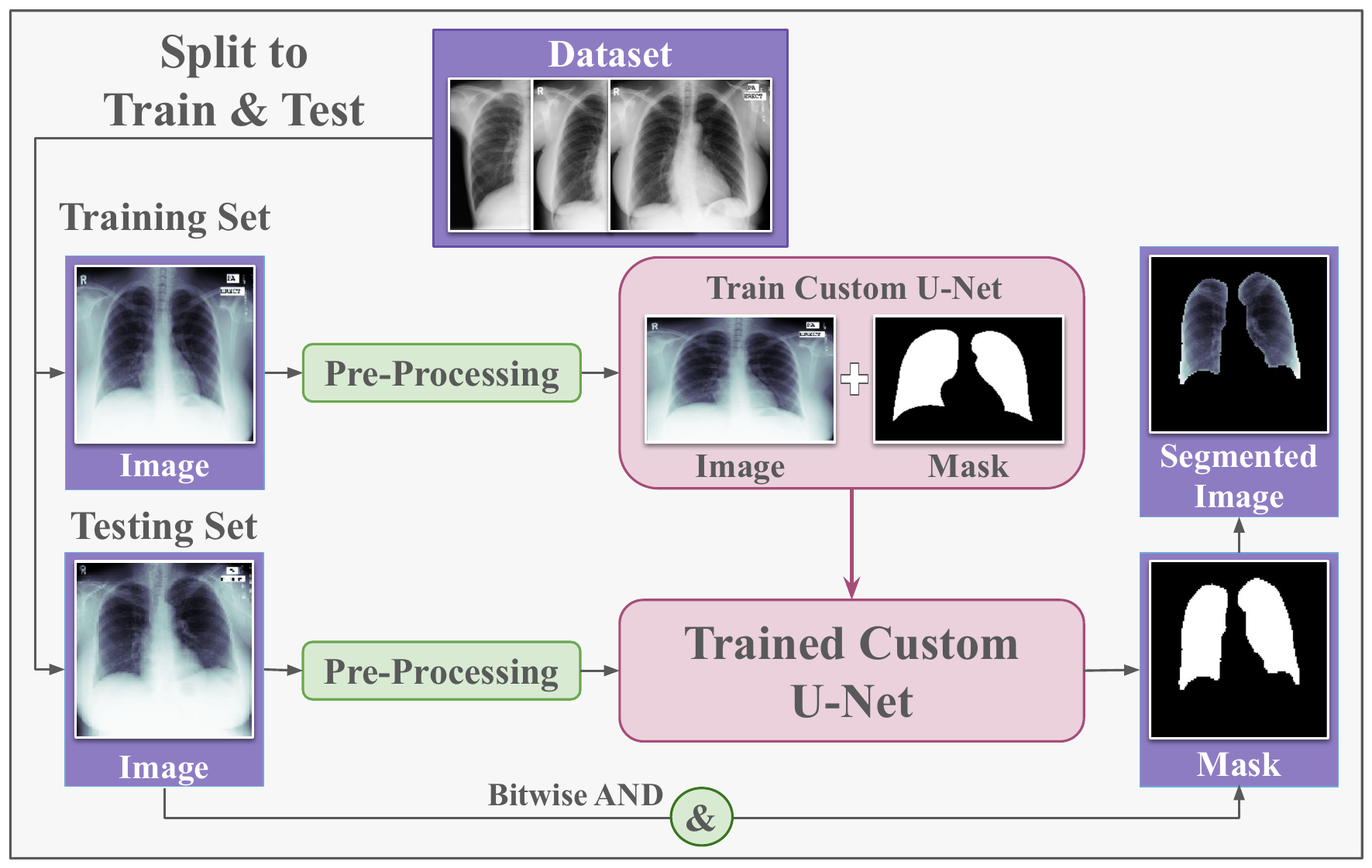}
    \caption{Segmentation using custom U-Net.}
    \label{2}
\end{figure}
\vspace{-0.3cm}
Segmentation is performed using a trained U-Net to predict masks and segmented images for the entire chest X-ray dataset. The process of training and testing a U-Net for segmentation is shown in Figure \ref{2}. The chest X-ray dataset developed for screening of pulmonary diseases \cite{jaeger2014two} is used as the dataset also comes with the corresponding masks for training. The dataset is first divided into the training and testing sets, and in the training part, the X-ray images and mask are preprocessed using resizing and padding. Afterward, the custom U-Net is trained using this set of images and masks with batch normalization and sigmoid as output activation. Then, the trained model is tested using the image alone to generate a mask. The generated mask is thereby combined with the image to get the final segmented image.

\subsection{Training using ViT}
\vspace{-0.2cm}
\begin{figure}[h]
    \centering
    \includegraphics[scale=0.41]{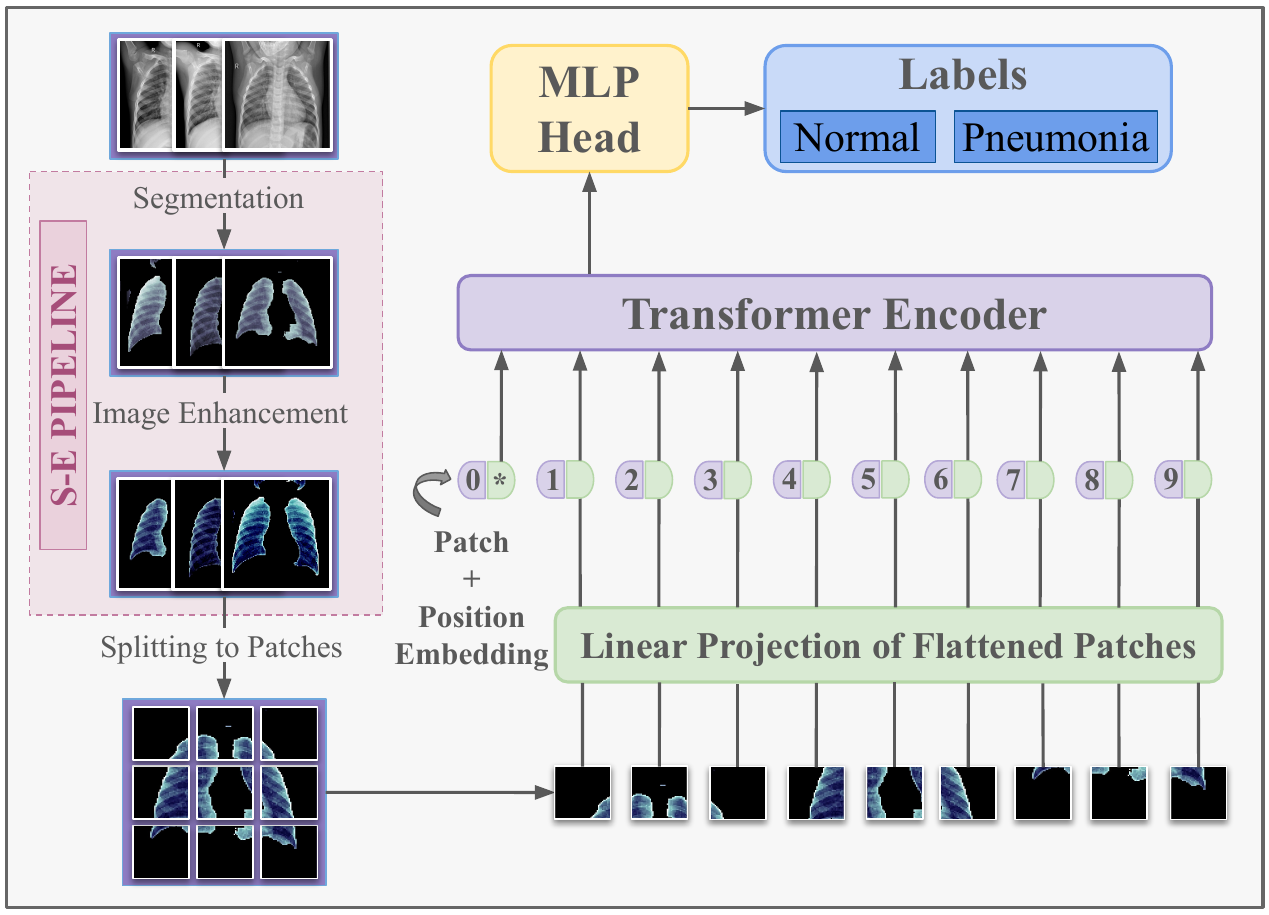}
    \caption{Workflow of the training process of the proposed defense mechanism in ViT.}
    \label{3}
\end{figure}
\vspace{-0.2cm}
 The training process of ViT for the proposed defense mechanism is depicted in Figure \ref{3}. Images from the dataset are first fed to the segmentation module, and then, any one of the image enhancement techniques, CLAHE, UM, or HFE is applied to the image. These two processes combine to form the pre-processing stage before feeding the input to ViT. The images are then divided into patches of equal size and all these patches are then fed to the linear embedding phase to convert it into a one-dimensional vector. After that, to each block, positional embedding is added to get the position of the patch in the original image. These positional embeddings are passed along with a special token for classification to the transformer encoder. The transformer encoder consists of a normalization layer, multi-head attention, residual connection, one more normalization layer, a multi-layer perception, and a residual connection. Finally, the MLP block does the classification job on two labels, normal and pneumonia, in our case.

\subsection{Calculating Rate of Reduction}
\begin{figure}
    \centering
    \includegraphics[scale=0.34]{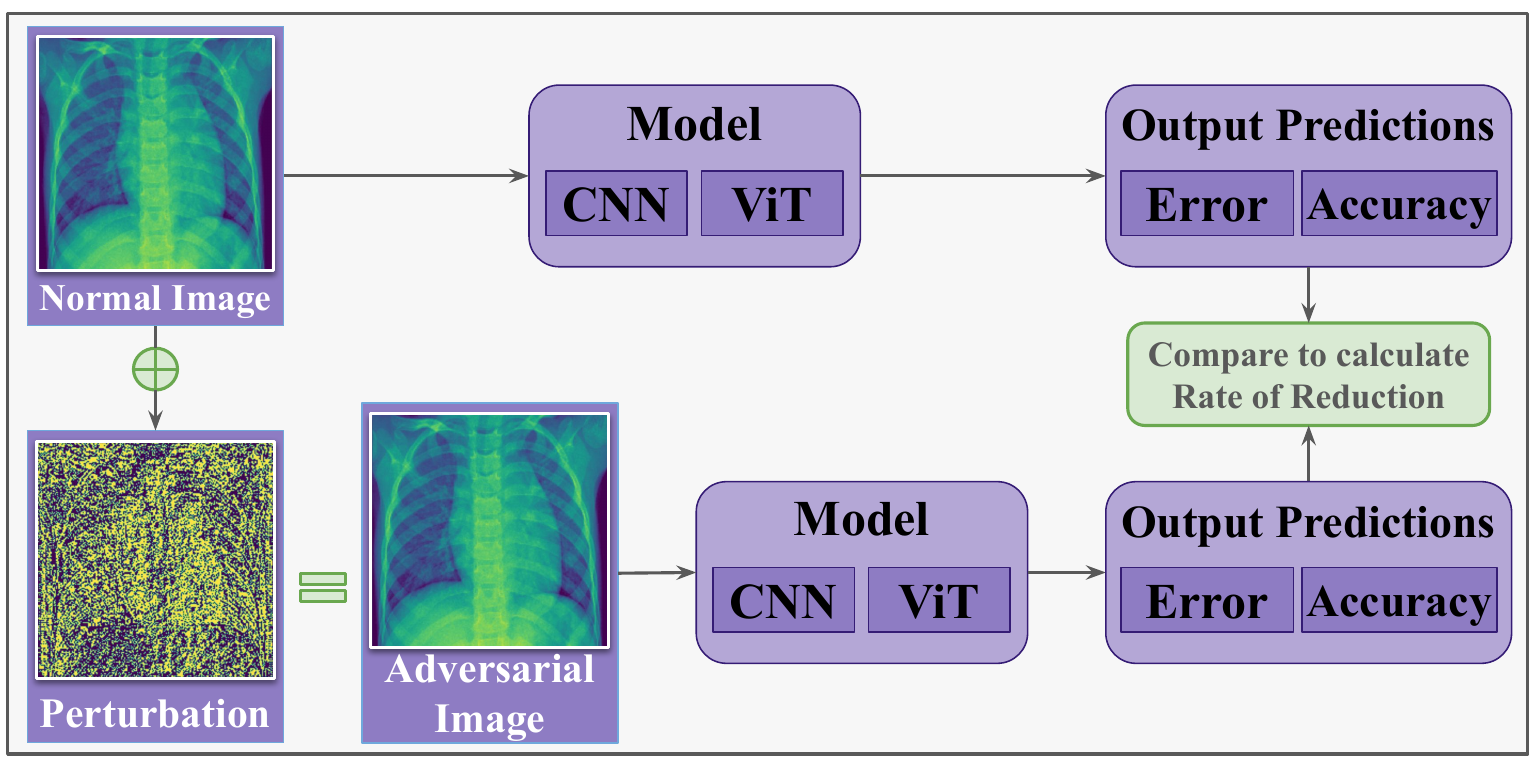}
    \caption{Comparsion of output predictions from normal and corresponding adversarial image.}
    \label{4}
\end{figure}
To calculate how much the effect of attack is reduced, we use the rate of reduction. The rate of reduction is calculated between the base model and our modified model with preprocessing. Adversarial attacks will always try to increase the probability of belonging to the incorrect class and decrease the probability of belonging to the correct class. The rate of reduction will give the measure of how much the attack is affecting by comparing two models. Formulas for calculating the rate of reduction are explained in Algorithm \ref{1}. For the base model, first, filter all examples where an attack is successful. Then, sum all the probabilities of the normal image belonging to class A and class B separately and find the average. Do a similar calculation for adversarial images and find the difference between normal image calculation and adversarial image calculation. This process is depicted in Figure \ref{4}. This similar calculation is done for the pre-processing attached model and the difference is calculated as rate of reduction.

\section{Experimental setup}
\subsection{Model Specifications and Architectures}
The details about the models, CNN and ViT, used in our experiments are shown in Table \ref{model}. The CNN model used is written from scratch and uses the sigmoid activation function in the last layer. ViT models used are vit\_b32 and vit\_l32, rectified adam as the optimizer, and uses early stopping technique by monitoring validation accuracy. 
\begin{table}[h]
\caption{Details of models used in our experiments.}
\begin{center}
\renewcommand{\arraystretch}{1.3}
\resizebox{\columnwidth}{!}{%
\begin{tabular}{||c | c | c |c | c | c ||} 
 \hline
\multirow{2}{4em}{Model} & \multicolumn{3}{|c|}{{ViT backbone}} & \multirow{2}{4em}{Parameters} & \multirow{2}{4em}{Accuracy} \\  \cline{2-4}
&  Layers & Hidden size & MLP size &  & \\
 \hline\hline
ViT-B-32 & 12 & 768 & 3072 & 86M & 90\% \\ 
 \hline
ViT-L-32 & 24 & 1024 & 4096  & 307M &  91\%\\
\hline \hline
\multicolumn{6}{c}{\hspace{-1cm}CNN convolutional layers} \\
\hline \hline
Simple CNN & \multicolumn{3}{c|}{12} & 25M & 94\% \\ 
\hline \hline
\end{tabular}
}
\end{center}
\label{model}
\end{table}

\subsection{Adversarial Attacks}
We consider two white-box adversarial attacks in our study, FGSM and PGD. For the FGSM attack, we create adversarial perturbation using the derivative of the loss function, and the attack is created for different $\epsilon$ values. PGD, which is the modified iterative version of FGSM where the attack strength is limited, is also implemented. 
\subsection{Image Enhancement Techniques}
\subsubsection{CLAHE}
CLAHE algorithm is a special type of histogram equalization that is used commonly to get high-contrast images. The image is split into tiles and then the algorithm works on each tile. Bilinear interpolation is used to blend the boundaries between neighboring tiles. The contrast is limited by clipping the histogram to a particular value which is predefined. 

\subsubsection{Unsharp Masking}
Unsharp Masking is used to sharpen the edges of an image by subtracting an unsharpened/blurred image. The algorithm consists of two steps: construct a blurred version of the image, and second, subtract the unsharpened image from the original image.

\begin{multline}
    sharpened\_image = original\_image + \\
    amount * (unsharpened\_mask)
\end{multline}


where the amount controls the edge intensity. The blurred image can be formed using different blurs, gaussian, median, maximum, and minimum.

\subsubsection{High-frequency Emphasis filtering}
High-frequency emphasis filter is also an image enhancement technique to sharpen the edges. Histogram equalization is used along with the technique to improve the contrast of the image as the output of this method can be low contrast. 

\begin{multline}
    sharpened\_image = (original\_image + \\
    (Gaussian\_Highpass\_Filter)) * (Histogram\_Equalization)
\end{multline}


\section{Experiments and Results}
Extensive experiments are performed using different image enhancement techniques and segmentation to study the effect of each technique and combined on different models. Image enhancement techniques are also implemented with different parameters to study the effects.
First, we did experiments using CNN with the S-E pipeline and got promising results. Then, we did the same set of experiments for ViTs and the results are discussed in the coming subsections.

\subsection{Performance of defense mechanism in CNN}
Tables \ref{fgsmcnn} and \ref{pgdcnn} show the rate of reduction for FGSM and PGD attacks respectively for various parameter settings. The best results for each set of enhancement techniques are highlighted in bold. For all the techniques except CLAHE, the best result is shown after segmentation.
\begin{table}[h]
\caption{Rate of reduction for FGSM attack in CNN for epsilon = 0.001.}
\begin{center}
\renewcommand{\arraystretch}{1.3}
\resizebox{\columnwidth}{!}{%
\begin{tabular}{||c | c | c |c | c||} 
 \hline
 \multicolumn{3}{||c|}{ \multirow{2}{13em}{ Image enhancement technique} } & {{Without segmentation}} & {With segmentation}\\ 
 \cline{4-5}
 \multicolumn{3}{||c|}{} &  Rate of reduction (in \%)  & Rate of reduction (in \%) \\
 \hline \hline
 \multicolumn{3}{||c|}{CLAHE} & \textbf{82.75} & 72.08
\\
 \hline
 \multirow{9}{6em}{UM-Gaussian} & \multirow{3}{4em}{Radius 5} & Amount 2 & 2.83 & 73.05\\
 \cline{3-5}
 & & Amount 3 & 44.70 & 72.84\\
 \cline{3-5}
 & & Amount 4 & 61.26 & 74.47\\
 \cline{2-5}
 & \multirow{3}{4em}{Radius 6} & Amount 2 & 51.49 & 70.30\\
 \cline{3-5}
 & & Amount 3 & 0 & 74.55\\
 \cline{3-5}
 & & Amount 4 & 18.88 & 74.02\\
 \cline{2-5}
  & \multirow{3}{4em}{Radius 7} & Amount 2 & 0 & 62.69\\
 \cline{3-5}
 & & Amount 3 & 10.89 & \textbf{77.61}\\
 \cline{3-5}
 & & Amount 4 & 24.25 & 66.87\\
 \cline{1-5}

 \multirow{3}{6em}{UM-Median} & \multicolumn{2}{c|}{Amount 2} & 0 & 71.64\\
 \cline{2-5}
 & \multicolumn{2}{c|}{Amount 3} & 19.40 & 73.73\\
 \cline{2-5}
 & \multicolumn{2}{c|}{Amount 4} & 22.26 & \textbf{77.31}\\
 \cline{1-5}

 \multirow{3}{7em}{UM-Maximum} & \multicolumn{2}{c|}{Amount 2} & 0 & 60.30\\
 \cline{2-5}
 & \multicolumn{2}{c|}{Amount 3} & 6.87 & 34.33\\
 \cline{2-5}
 & \multicolumn{2}{c|}{Amount 4} & 8.89 & 27.76\\
 \cline{1-5}

 \multirow{3}{7em}{UM-Minimum} & \multicolumn{2}{c|}{Amount 2} & 37.74 & 84.40\\
 \cline{2-5}
 & \multicolumn{2}{c|}{Amount 3} & 60.46 & 80.00\\
 \cline{2-5}
 & \multicolumn{2}{c|}{Amount 4} & 77.24 & \textbf{84.70}\\
 \cline{1-5}

 \multirow{5}{2em}{HFE} & \multicolumn{2}{c|}{10} & 0 & \textbf{80.07}\\
 \cline{2-5}
 & \multicolumn{2}{c|}{20} & 5.07 & 48.88\\
 \cline{2-5}
 & \multicolumn{2}{c|}{30} & 0 & 67.91\\
 \cline{2-5}
 & \multicolumn{2}{c|}{40} & 0 & 67.87\\
 \cline{2-5}
 & \multicolumn{2}{c|}{50} & 42.04 & 54.69\\
 \cline{1-5}
 
\end{tabular}
}
\end{center}
\label{fgsmcnn}
\end{table}

\begin{table}
\caption{Rate of reduction for PGD attack in CNN for epsilon = 1/255, step size = 2 and iterations = 7.}
\begin{center}
\renewcommand{\arraystretch}{1.3}
\resizebox{\columnwidth}{!}{%
\begin{tabular}{||c | c | c |c | c||} 
 \hline
 \multicolumn{3}{||c|}{ \multirow{2}{13em}{ Image enhancement technique } } & {{Without segmentation}} & {With segmentation}\\ 
 \cline{4-5}
 \multicolumn{3}{||c|}{} &  Rate of reduction (in \%) & Rate of reduction (in \%) \\
 \hline \hline
 \multicolumn{3}{||c|}{CLAHE} & \textbf{73.44} & 50.11 \\
 \hline
 \multirow{9}{6em}{UM-Gaussian} & \multirow{3}{4em}{Radius 5} & Amount 2 & 47.25 & 71.14\\
 \cline{3-5}
 & & Amount 3 & 57.32 & 71.24\\
 \cline{3-5}
 & & Amount 4 & 57.32 & 71.24\\
 \cline{2-5}
 & \multirow{3}{4em}{Radius 6} & Amount 2 & 60.61 & 68.27\\
 \cline{3-5}
 & & Amount 3 & 47.46 & 73.34\\
 \cline{3-5}
 & & Amount 4 & 44.97 & 70.47\\
 \cline{2-5}
  & \multirow{3}{4em}{Radius 7} & Amount 2 & 48.35 & 63.93\\
 \cline{3-5}
 & & Amount 3 & 46.63 & \textbf{75.42}\\
 \cline{3-5}
 & & Amount 4 & 45.93 & 63.9\\
 \cline{1-5}

 \multirow{3}{6em}{UM-Median} & \multicolumn{2}{c|}{Amount 2} & 44.58 & 72.35\\
 \cline{2-5}
 & \multicolumn{2}{c|}{Amount 3} & 49.21 & 73.02\\
 \cline{2-5}
 & \multicolumn{2}{c|}{Amount 4} & 49.76 & \textbf{75.77}\\
 \cline{1-5}

 \multirow{3}{7em}{UM-Maximum} & \multicolumn{2}{c|}{Amount 2} & 43.21 & \textbf{62.87}\\
 \cline{2-5}
 & \multicolumn{2}{c|}{Amount 3} & 45.19 & 50.33\\
 \cline{2-5}
 & \multicolumn{2}{c|}{Amount 4} & 48.57 & 52.05\\
 \cline{1-5}

 \multirow{3}{7em}{UM-Minimum} & \multicolumn{2}{c|}{Amount 2} & 50.46 & \textbf{82.79}\\
 \cline{2-5}
 & \multicolumn{2}{c|}{Amount 3} & 63.32 & 75.93\\
 \cline{2-5}
 & \multicolumn{2}{c|}{Amount 4} & 75.96 & 82.70\\
 \cline{1-5}

 \multirow{5}{2em}{HFE} & \multicolumn{2}{c|}{10} & 47.33 & \textbf{82.09}\\
 \cline{2-5}
 & \multicolumn{2}{c|}{20} & 45.70 & 58.72\\
 \cline{2-5}
 & \multicolumn{2}{c|}{30} & 44.46 & 68.81\\
 \cline{2-5}
 & \multicolumn{2}{c|}{40} & 44.07 & 70.73\\
 \cline{2-5}
 & \multicolumn{2}{c|}{50} & 57.74 & 59.46\\
 \cline{1-5}
 
\end{tabular}
}
\end{center}
\label{pgdcnn}
\end{table}

\begin{itemize}
    \item \textbf{CLAHE:} The effect of CLAHE image enhancement is studied in CNN and the rate of reduction for various epsilons from 0.001 to 0.005 is depicted in Figure \ref{clahecnn}. The rate of reduction is calculated by using the amount of change an attack can make in a model and comparing it with the base model as explained in Algorithm \ref{alg}. The CLAHE enhancement was able to reduce the effect of attack by 82.75\% for epsilon value 0.001 and the rate of reduction value decreases as epsilon decreases. As epsilon increases, more stronger attacks will be performed and thus will be difficult to counter. However, since stronger attacks will be perceptible to the human eye, an attacker will not choose a larger epsilon value and thus the defense mechanism performs well in such scenarios. 
    \vspace{-0.3cm}
    \begin{figure}[ht]

  \subcaptionbox{Without segmentation}[.5\linewidth]{%
    \includegraphics[width=\linewidth]{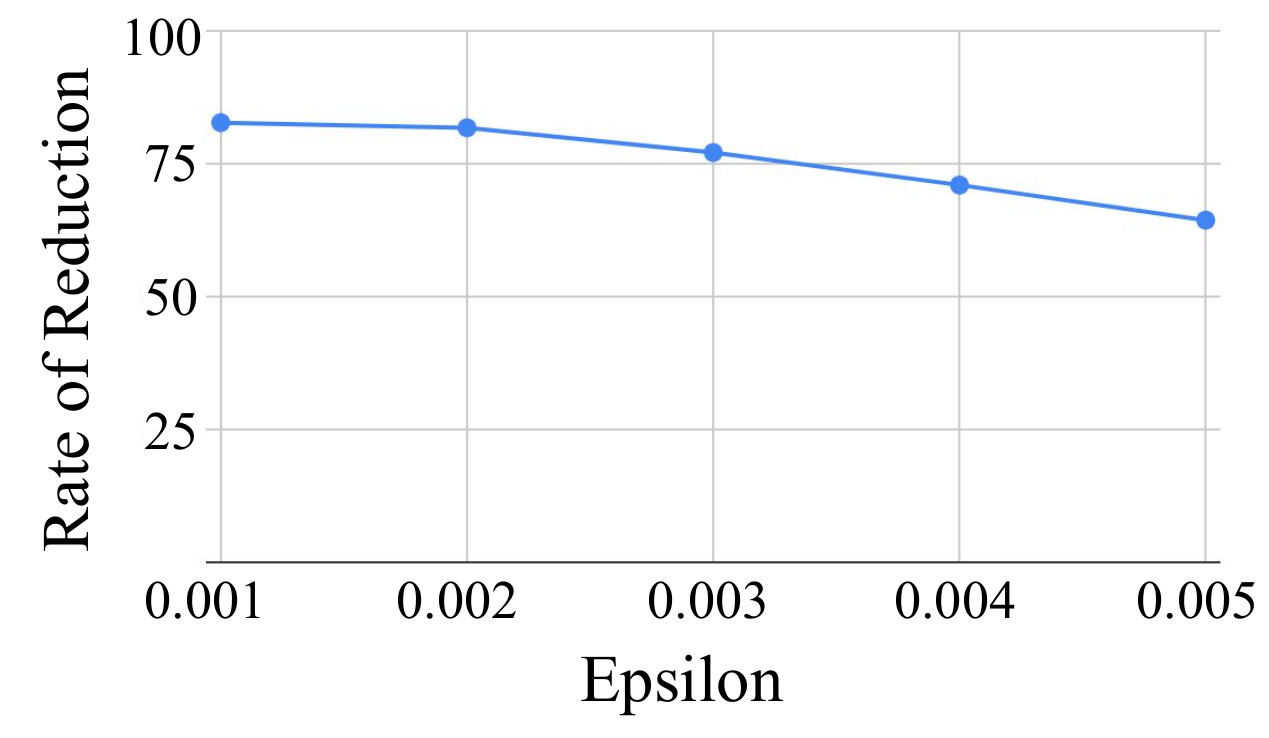}%
  }%
  \hfill
  \subcaptionbox{With segmentation}[.5\linewidth]{%
    \includegraphics[width=\linewidth]{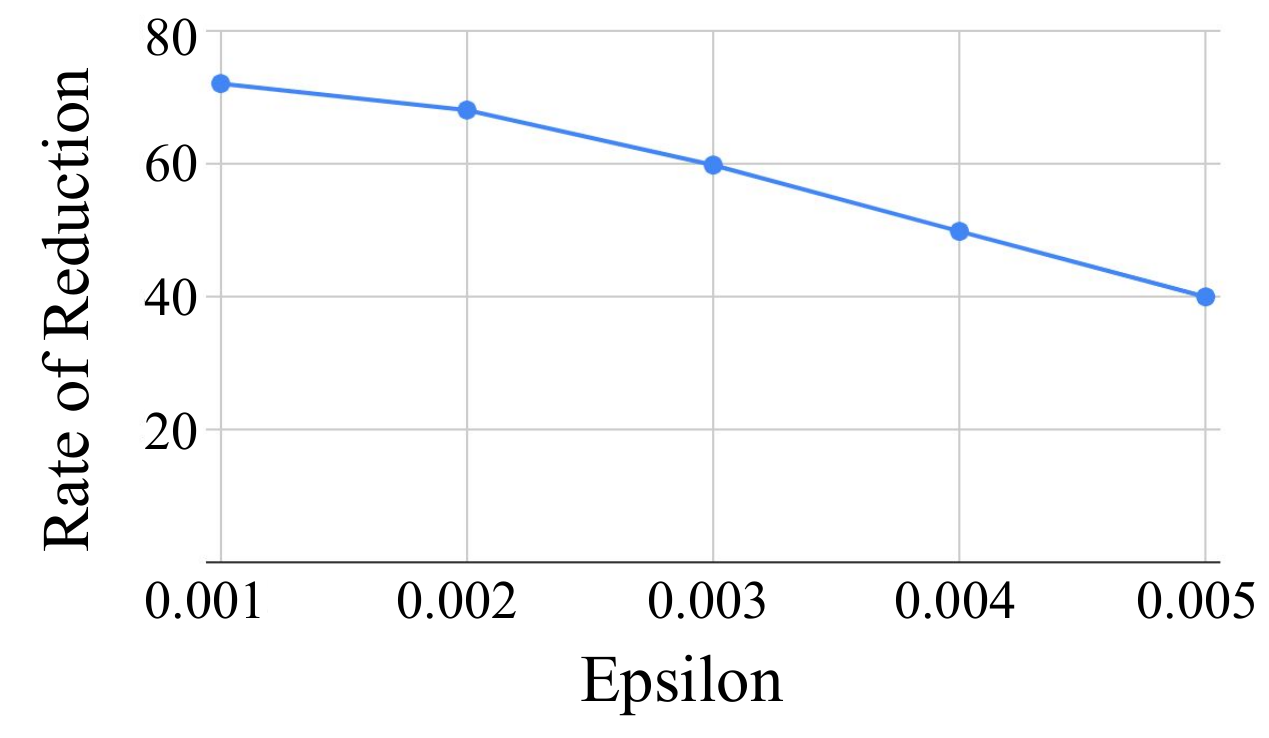}%
  }
  \caption{Rate of reduction for CNN using CLAHE after performing FGSM attack for epsilon values from 0.001 to 0.005.}
  \label{clahecnn}
\end{figure}

    \item \textbf{Unsharp Masking:} 
    Figures \ref{gaussiancnn} to \ref{minimumcnn} demonstrate the rate of reduction for different filters, gaussian, median, maximum, and minimum. As discussed in Figure \ref{clahecnn}, the figure shows the rate of reduction for different epsilon values. 
    Figures \ref{gaussiancnn} to \ref{minimumcnn} also show the rate of reduction for different filters, gaussian, median, maximum, and minimum for segmented images. The images are first segmented and then enhanced using any of the filters to get the corresponding results. As we can see from the graphs, the rate of reduction has improved significantly when segmentation is included. For selecting the best parameter setting, Figure \ref{heatcnn} shows the heatmap for the rate of reduction values for various parameters of the Gaussian filter.
    \vspace{-0.2cm}
    \begin{figure}[h]
  \subcaptionbox{}[.24\linewidth]{%
    \includegraphics[width=\linewidth]{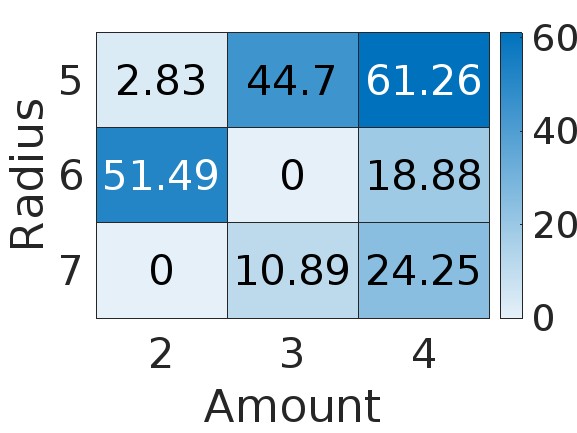}%
  }%
  \hfill
  \subcaptionbox{}[.24\linewidth]{%
    \includegraphics[width=\linewidth]{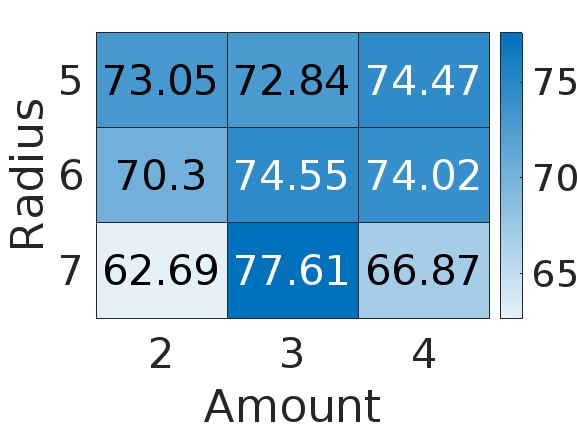}%
  }
  \subcaptionbox{}[.24\linewidth]{%
    \includegraphics[width=\linewidth]{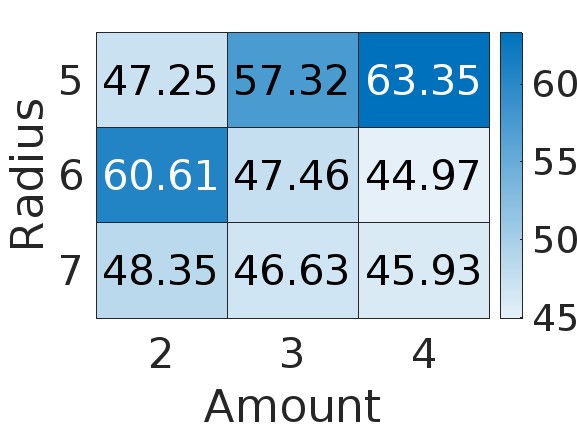}%
  }%
  \hfill
  \subcaptionbox{}[.24\linewidth]{%
    \includegraphics[width=\linewidth]{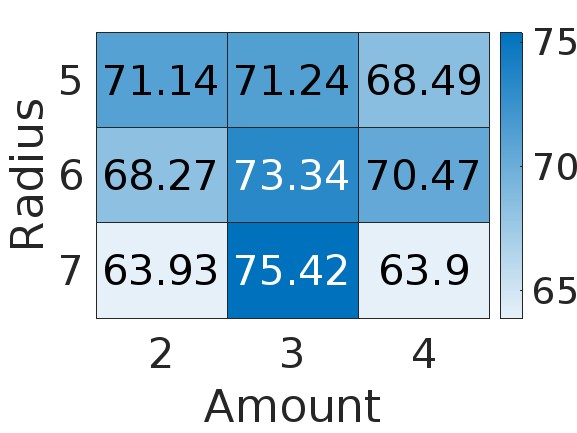}%
  }
  \caption{Rate of reduction for FGSM attack ((a)Without segmentation, (b)With segmentation) \& PGD attack ((c)Without segmentation, (d)With segmentation) for CNN using Gaussian filter for different parameter values, amount, \& radius.}
  \label{heatcnn}
\end{figure}

\vspace{-0.6cm}

\begin{figure}[h]
  \subcaptionbox{Without segmentation}[.5\linewidth]{%
    \includegraphics[width=\linewidth]{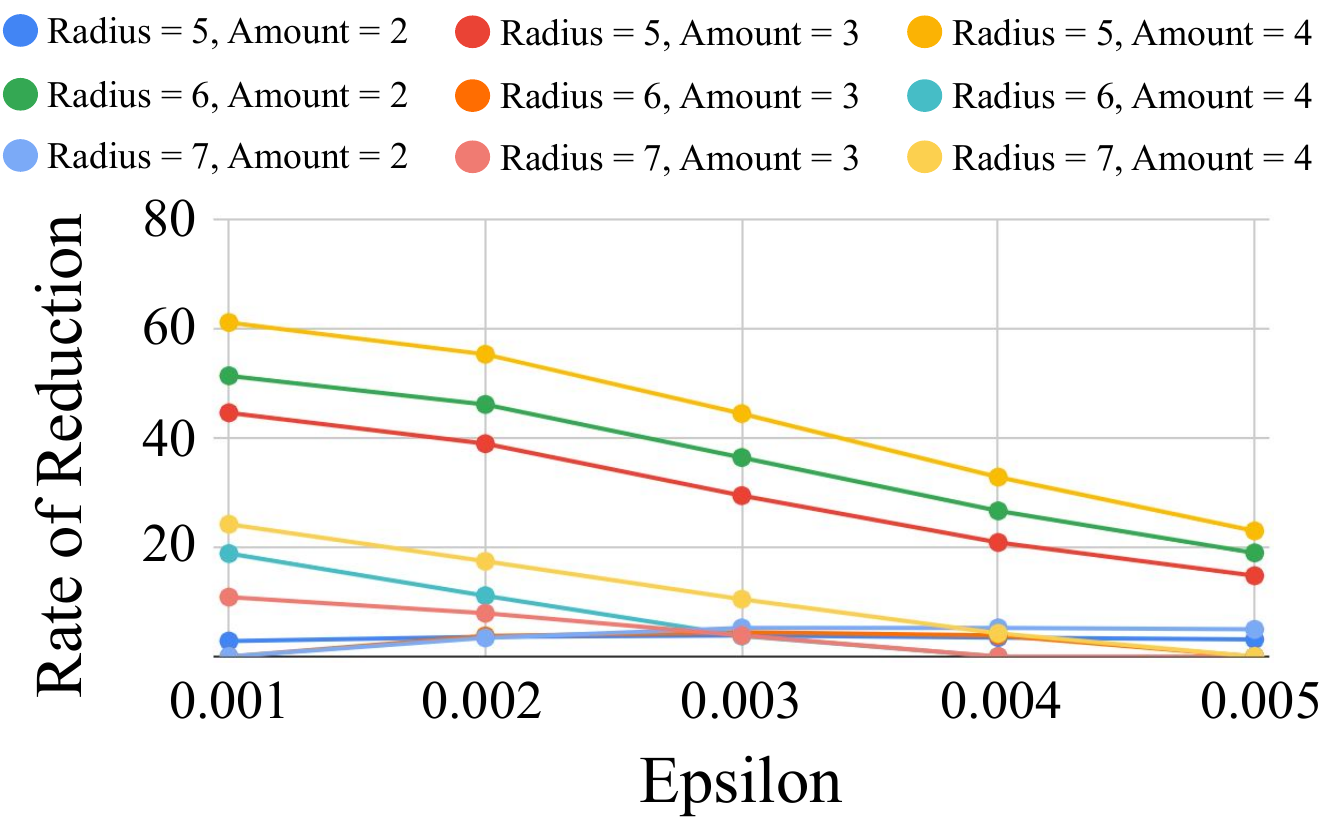}%
  }%
  \hfill
  \subcaptionbox{With segmentation}[.5\linewidth]{%
    \includegraphics[width=\linewidth]{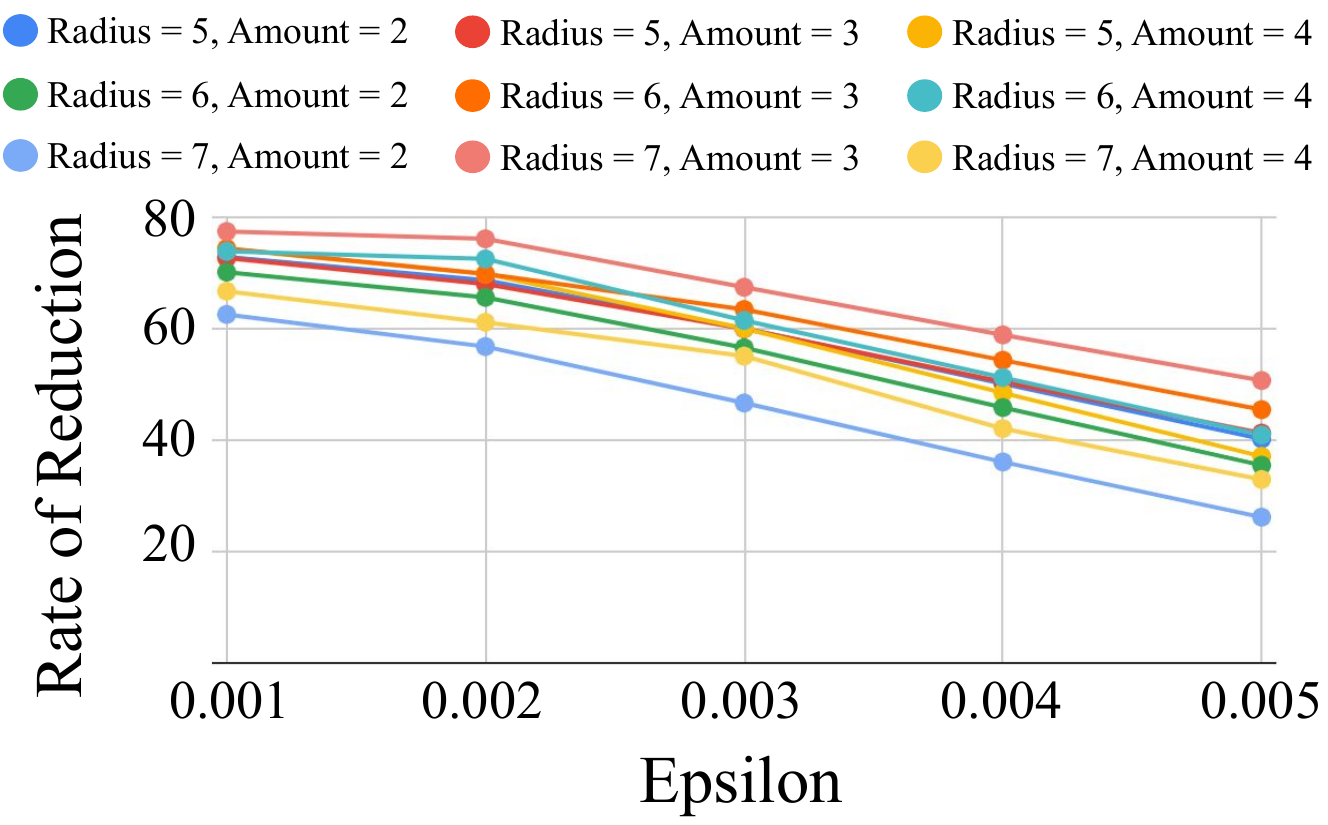}%
  }
  \caption{Rate of reduction for CNN using UM-Gaussian after performing FGSM attack for epsilon values: 0.001 to 0.005.}
  \label{gaussiancnn}
\end{figure}

\vspace{-0.6cm}

\begin{figure}[ht]
  \subcaptionbox{Without segmentation}[.5\linewidth]{%
    \includegraphics[width=\linewidth]{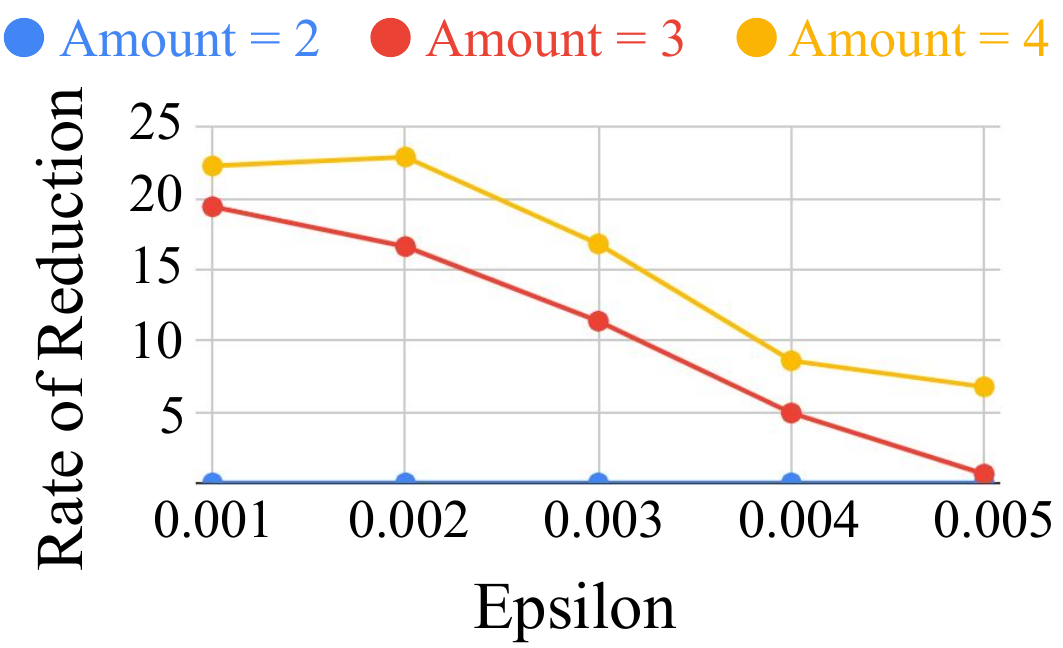}%
  }%
  \hfill
  \subcaptionbox{With segmentation}[.5\linewidth]{%
    \includegraphics[width=\linewidth]{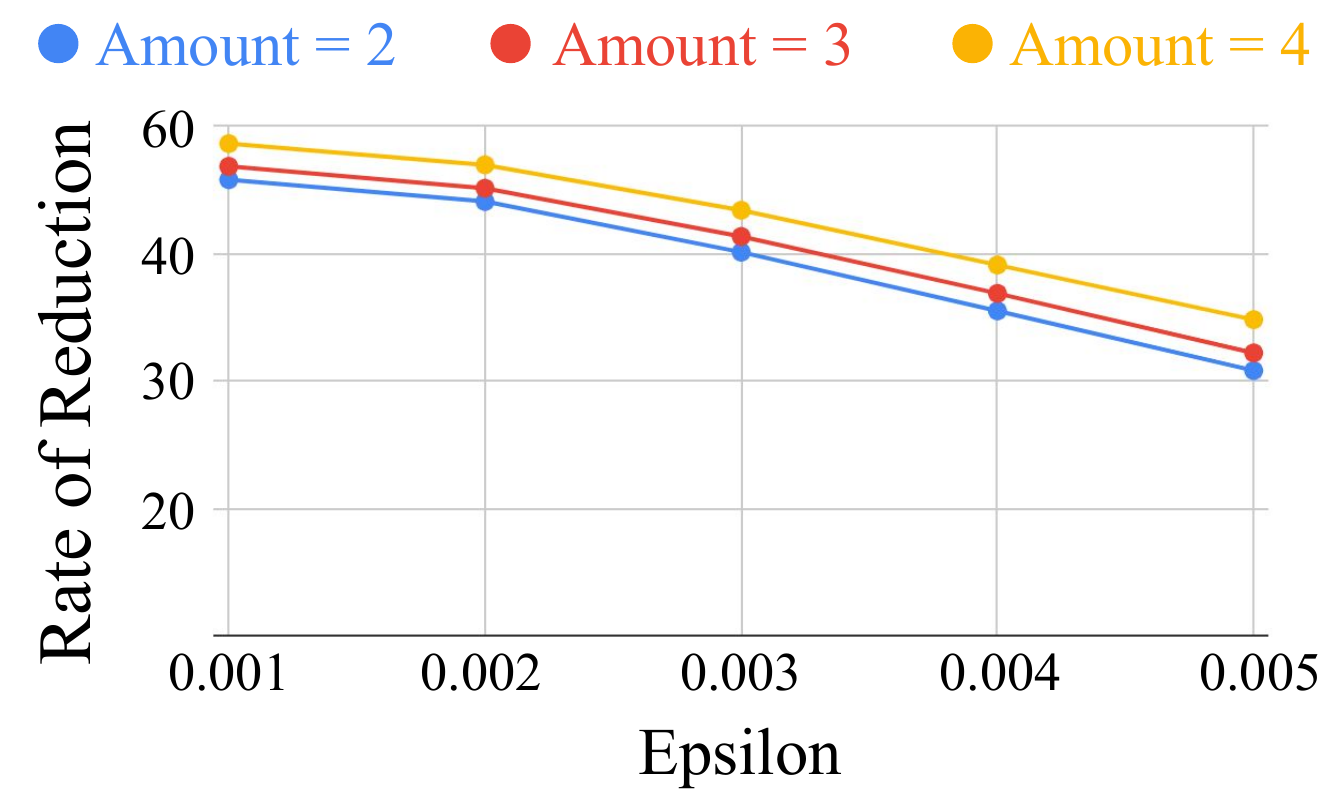}%
  }
  \caption{Rate of reduction for CNN using UM-Median after performing FGSM attack for epsilon values: 0.001 to 0.005.}
\end{figure}

\vspace{-0.6cm}

\begin{figure}[h]
  \subcaptionbox{Without segmentation}[.5\linewidth]{%
    \includegraphics[width=\linewidth]{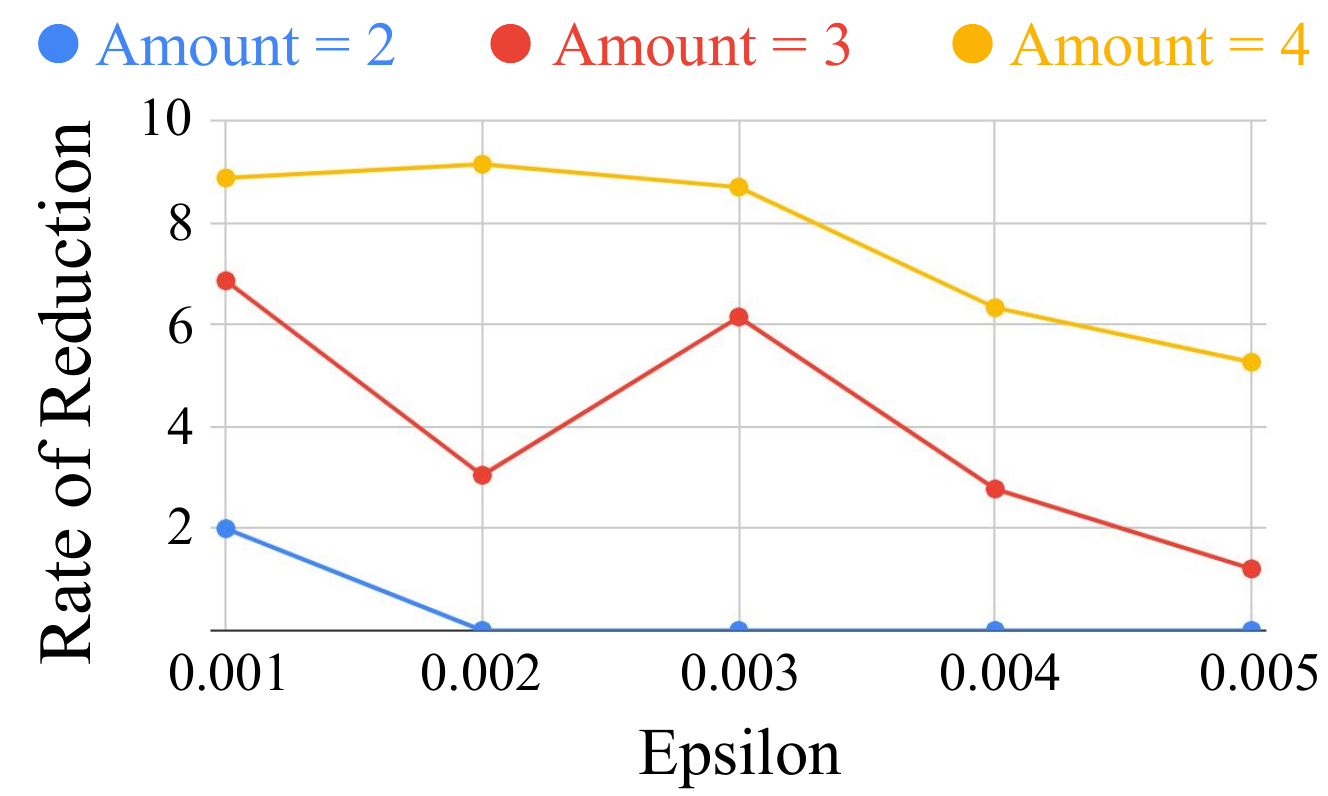}%
  }%
  \hfill
  \subcaptionbox{With segmentation}[.5\linewidth]{%
    \includegraphics[width=\linewidth]{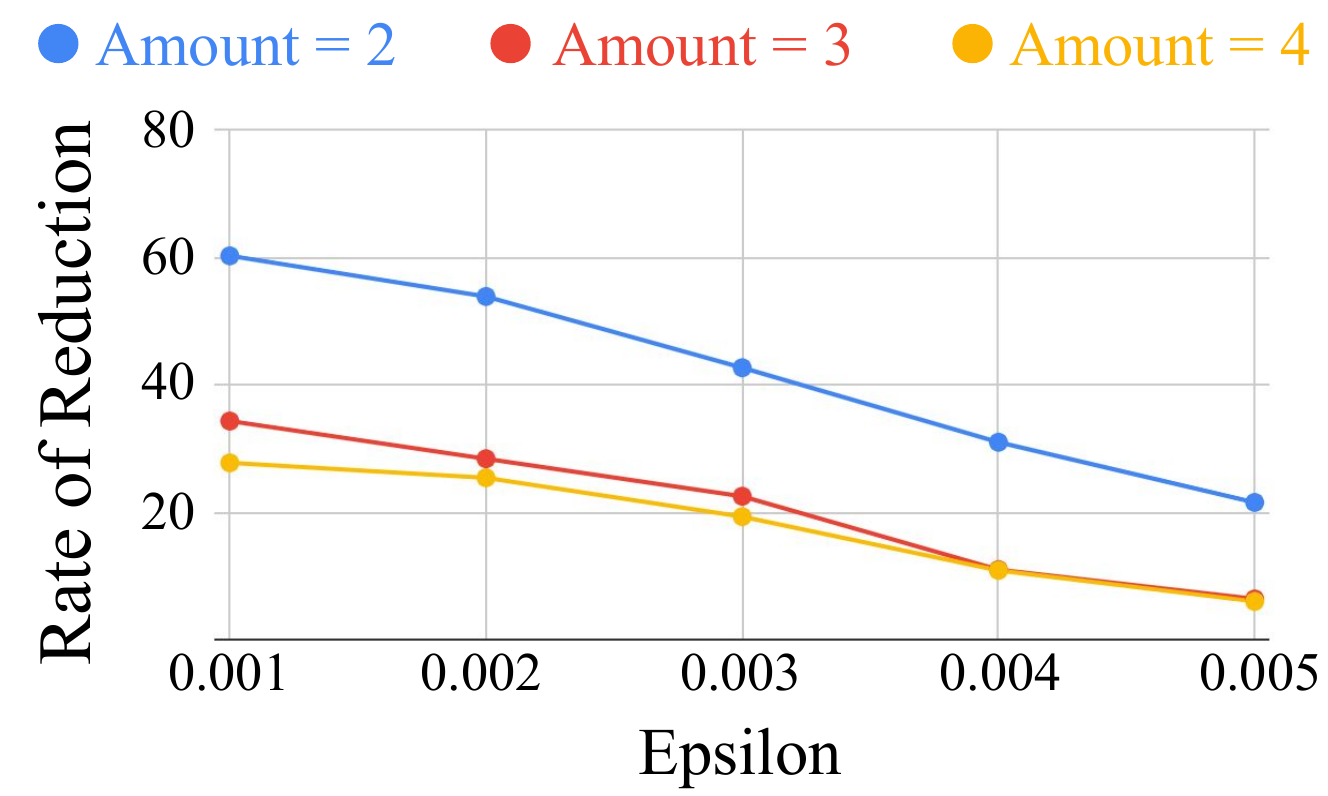}%
  }
  \caption{Rate of reduction for CNN using UM-Maximum after performing FGSM attack for epsilon values: 0.001 to 0.005.}
\end{figure}

\begin{figure}[h]
  \subcaptionbox{Without segmentation}[.5\linewidth]{%
    \includegraphics[width=\linewidth]{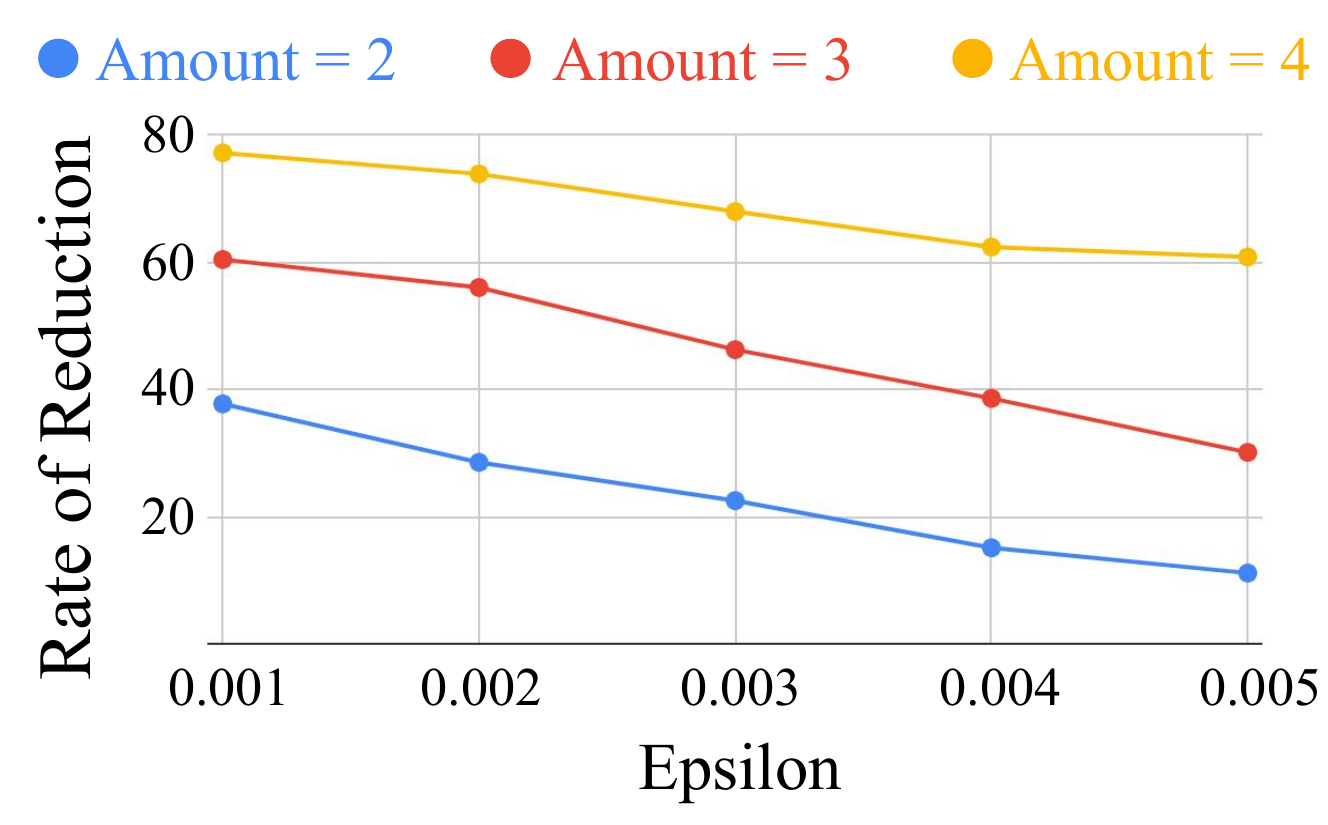}%
  }%
  \hfill
  \subcaptionbox{With segmentation}[.5\linewidth]{%
    \includegraphics[width=\linewidth]{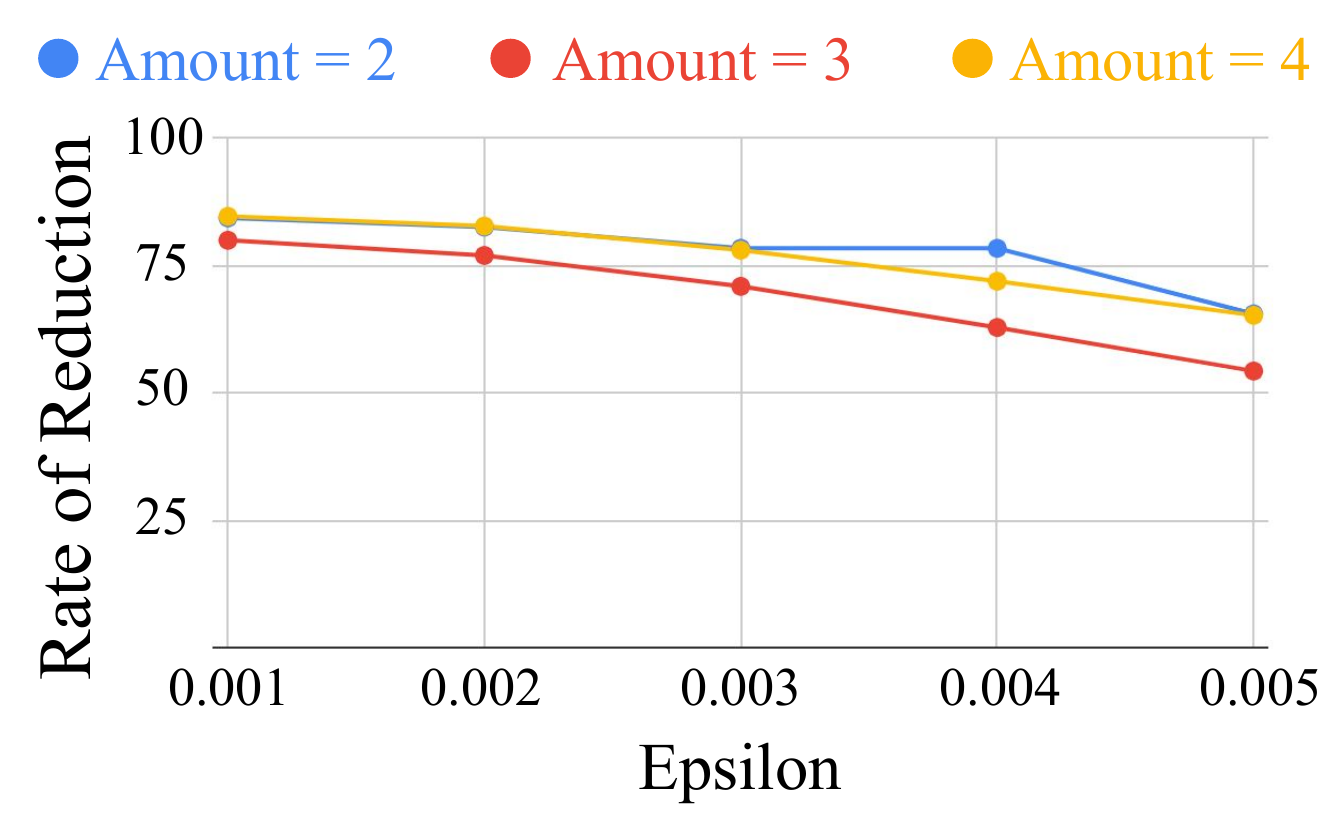}%
  }
  \caption{Rate of reduction for CNN using UM-Minimum after performing FGSM attack for epsilon values: 0.001 to 0.005.}
  \label{minimumcnn}
\end{figure}

\item \textbf{High-Frequency Emphasis filtering:}
Results for high-frequency emphasis filtering are demonstrated in Figure \ref{hfecnn}. Similar to the above results, segmentation adds to the rate of reduction for all values considered from 10 to 50, and this value acts as the distance from the center of the Fourier image, which is shifted.
\begin{figure}[ht]
  \subcaptionbox{Without segmentation}[.5\linewidth]{%
    \includegraphics[width=\linewidth]{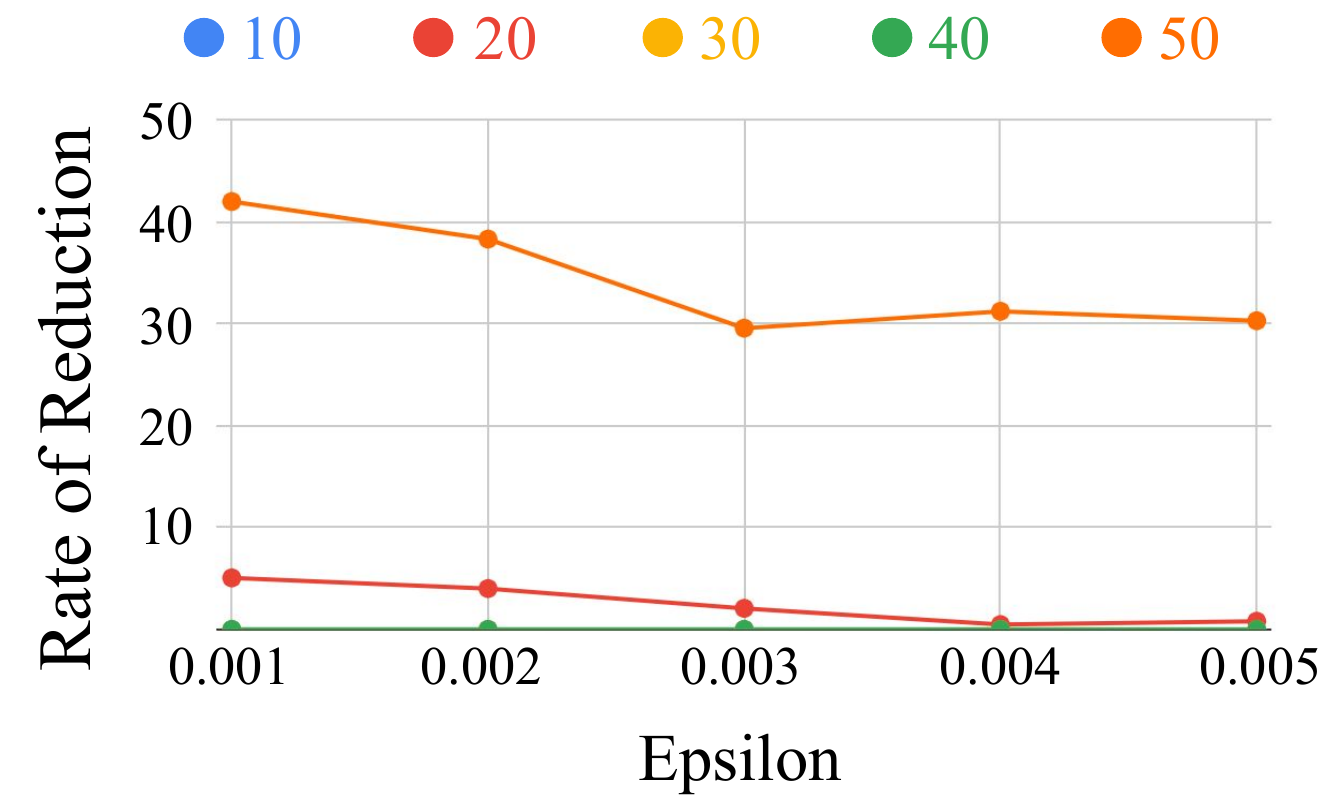}%
  }%
  \hfill
  \subcaptionbox{With segmentation}[.5\linewidth]{%
    \includegraphics[width=\linewidth]{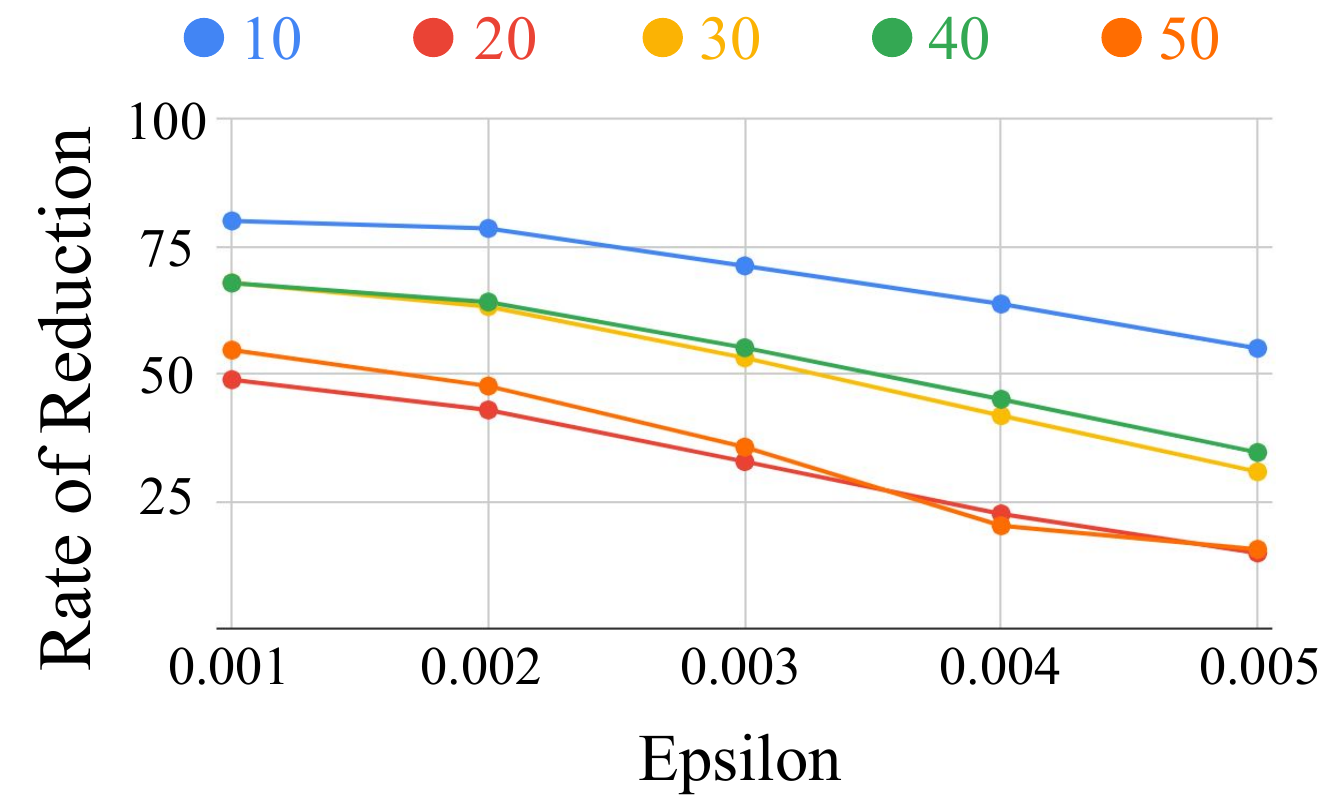}%
  }
  \caption{Rate of reduction for CNN using HFE after performing FGSM attack for epsilon values from 0.001 to 0.005.}
  \label{hfecnn}
\end{figure}

\end{itemize}

\subsection{Performance of defense mechanism in ViT-b32}

\begin{table}
\vspace{0.1cm}
\caption{Rate of reduction for FGSM attack in ViT-b32 for epsilon value = 0.001.}
\begin{center}
\renewcommand{\arraystretch}{1.3}
\resizebox{\columnwidth}{!}{%
\begin{tabular}{||c | c | c |c | c||} 
 \hline
 \multicolumn{3}{||c|}{ \multirow{2}{13em}{ Image enhancement technique } } & {{Without segmentation}} & {With segmentation}\\ 
 \cline{4-5}
 \multicolumn{3}{||c|}{} &  Rate of reduction (in \%) & Rate of reduction (in \%)\\
 \hline \hline
 \multicolumn{3}{||c|}{CLAHE} & 28.46 & \textbf{42.67}\\
 \hline
 \multirow{9}{6em}{UM-Gaussian} & \multirow{3}{4em}{Radius 5} & Amount 2 & 48.48 & 54.82\\
 \cline{3-5}
 & & Amount 3 & 10.14 & 50.08\\
 \cline{3-5}
 & & Amount 4 & 18.15 & \textbf{67.74}\\
 \cline{2-5}
 & \multirow{3}{4em}{Radius 6} & Amount 2 & 50.45 & 37.72\\
 \cline{3-5}
 & & Amount 3 & 32.38 & 37.72\\
 \cline{3-5}
 & & Amount 4 & 22.71 & 59.63\\
 \cline{2-5}
  & \multirow{3}{4em}{Radius 7} & Amount 2 & 17.80 & 42.72\\
 \cline{3-5}
 & & Amount 3 & 49.57 & 46.54\\
 \cline{3-5}
 & & Amount 4 & 31.29 & 47.60\\
 \cline{1-5}

 \multirow{3}{6em}{UM-Median} & \multicolumn{2}{c|}{Amount 2} & 17.41 & 35.97\\
 \cline{2-5}
 & \multicolumn{2}{c|}{Amount 3} & 21.29 & 48.71\\
 \cline{2-5}
 & \multicolumn{2}{c|}{Amount 4} & 13.82 & \textbf{66.57}\\
 \cline{1-5}

 \multirow{3}{7em}{UM-Maximum} & \multicolumn{2}{c|}{Amount 2} & 22.42 & 30.70\\
 \cline{2-5}
 & \multicolumn{2}{c|}{Amount 3} & 6.68 & 32.61\\
 \cline{2-5}
 & \multicolumn{2}{c|}{Amount 4} & 0 & \textbf{33.38}\\
 \cline{1-5}

 \multirow{3}{7em}{UM-Minimum} & \multicolumn{2}{c|}{Amount 2} & 54.35 & 26.38\\
 \cline{2-5}
 & \multicolumn{2}{c|}{Amount 3} & \textbf{72.22} & 47.95\\
 \cline{2-5}
 & \multicolumn{2}{c|}{Amount 4} & 46.01 & 58.63\\
 \cline{1-5}

 \multirow{5}{2em}{HFE} & \multicolumn{2}{c|}{10} & 32.82 & 30.49\\
 \cline{2-5}
 & \multicolumn{2}{c|}{20} & 19.23 & 20.87\\
 \cline{2-5}
 & \multicolumn{2}{c|}{30} & 43.42 & 32.99\\
 \cline{2-5}
 & \multicolumn{2}{c|}{40} & 25.27 & 43.99\\
 \cline{2-5}
 & \multicolumn{2}{c|}{50} & 19.78 & \textbf{53.67}\\
 \cline{1-5}
 
\end{tabular}
}
\end{center}
\label{fgsmvit}
\end{table}

\begin{table}

\caption{Rate of reduction for PGD attack in ViT-b32 for epsilon = 1/255, step size = 2 and iterations = 7.}
\begin{center}
\renewcommand{\arraystretch}{1.3}
\resizebox{\columnwidth}{!}{%
\begin{tabular}{||c | c | c |c | c||} 
 \hline
 \multicolumn{3}{||c|}{ \multirow{2}{13em}{Image enhancement technique } } & {{Without segmentation}} & {With segmentation}\\ 
 \cline{4-5}
 \multicolumn{3}{||c|}{} &  Rate of reduction (in \%)& Rate of reduction (in \%)\\
 \hline \hline
 \multicolumn{3}{||c|}{CLAHE} & 4.23 & \textbf{6.67}\\
 \hline
 \multirow{9}{6em}{UM-Gaussian} & \multirow{3}{4em}{Radius 5} & Amount 2 & 14.04 & 16.77\\
 \cline{3-5}
 & & Amount 3 & 0.5 & 7.79\\
 \cline{3-5}
 & & Amount 4 & 5.16 & \textbf{35.38}\\
 \cline{2-5}
 & \multirow{3}{4em}{Radius 6} & Amount 2 & 21.4 & 8.27\\
 \cline{3-5}
 & & Amount 3 & 8.54 & 8.2\\
 \cline{3-5}
 & & Amount 4 & 0.74 & 18.50\\
 \cline{2-5}
  & \multirow{3}{4em}{Radius 7} & Amount 2 & 0 & 7.25\\
 \cline{3-5}
 & & Amount 3 & 10.12 & 9.14\\
 \cline{3-5}
 & & Amount 4 & 2.93 & 15.09\\
 \cline{1-5}

 \multirow{3}{6em}{UM-Median} & \multicolumn{2}{c|}{Amount 2} & 2.22 & 2.02\\
 \cline{2-5}
 & \multicolumn{2}{c|}{Amount 3} & 2.63 & 29.30\\
 \cline{2-5}
 & \multicolumn{2}{c|}{Amount 4} & 2.97 & \textbf{34.74}\\
 \cline{1-5}

 \multirow{3}{7em}{UM-Maximum} & \multicolumn{2}{c|}{Amount 2} & 4.49 & 17.92\\
 \cline{2-5}
 & \multicolumn{2}{c|}{Amount 3} & 22.21 & 17.28\\
 \cline{2-5}
 & \multicolumn{2}{c|}{Amount 4} & 2.39 & \textbf{36.25}\\
 \cline{1-5}

 \multirow{3}{7em}{UM-Minimum} & \multicolumn{2}{c|}{Amount 2} & \textbf{19.88} & 5.46\\
 \cline{2-5}
 & \multicolumn{2}{c|}{Amount 3} & 5.53 & 9.55\\
 \cline{2-5}
 & \multicolumn{2}{c|}{Amount 4} & 1.35 & 16.64\\
 \cline{1-5}

 \multirow{5}{2em}{HFE} & \multicolumn{2}{c|}{10} & 18.90 & 9.48\\
 \cline{2-5}
 & \multicolumn{2}{c|}{20} & 11.98 & 9.89\\
 \cline{2-5}
 & \multicolumn{2}{c|}{30} & \textbf{19.07} & 13.87\\
 \cline{2-5}
 & \multicolumn{2}{c|}{40} & 15.86 & 14.75\\
 \cline{2-5}
 & \multicolumn{2}{c|}{50} & 8.27 & 17.25\\
 \cline{1-5}
 
\end{tabular}
}
\end{center}
\label{pgdvit}
\end{table}

\begin{figure}[ht]
  \subcaptionbox{Without segmentation}[.5\linewidth]{%
    \includegraphics[width=\linewidth]{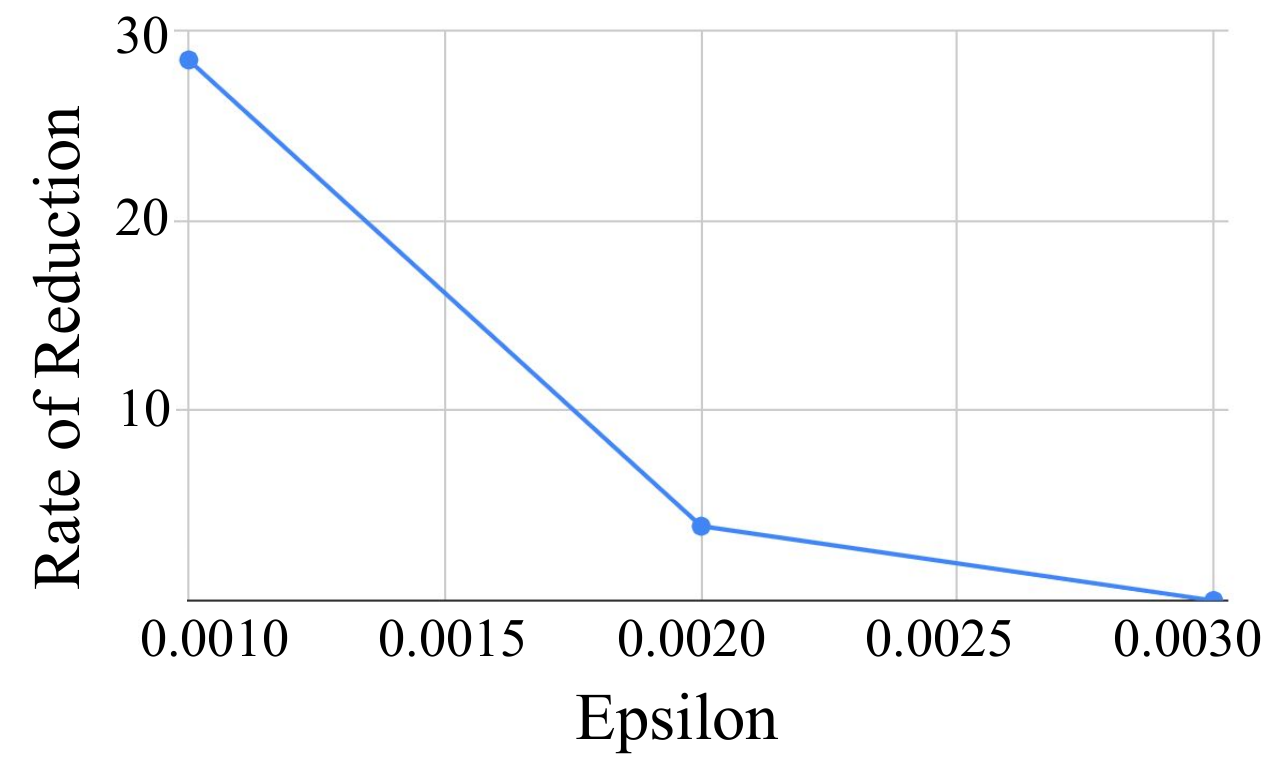}%
  }%
  \hfill
  \subcaptionbox{With segmentation}[.5\linewidth]{%
    \includegraphics[width=\linewidth]{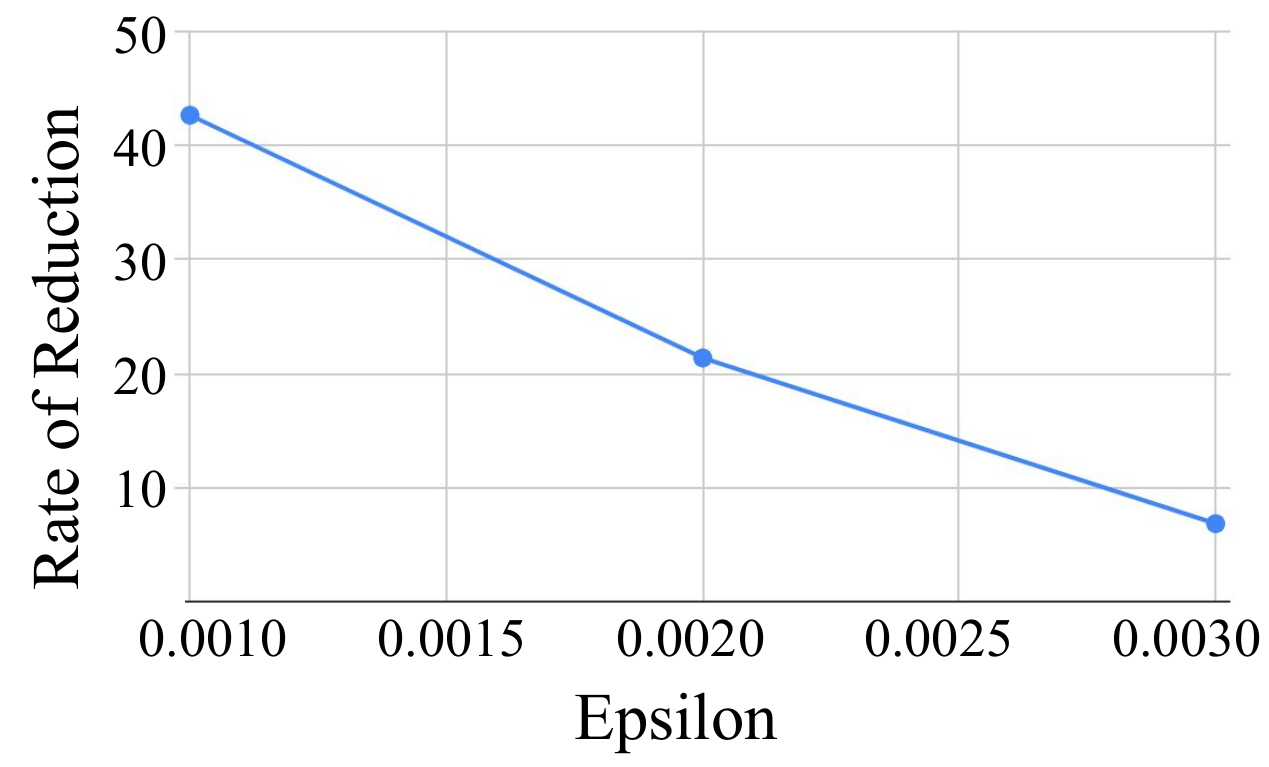}%
  }
  \caption{Rate of reduction for ViT using CLAHE after performing FGSM attack for epsilon values from 0.001 to 0.003.}
  \label{clahevit}
\end{figure}

\begin{figure}[ht]
  \subcaptionbox{Without segmentation}[.5\linewidth]{%
    \includegraphics[width=\linewidth]{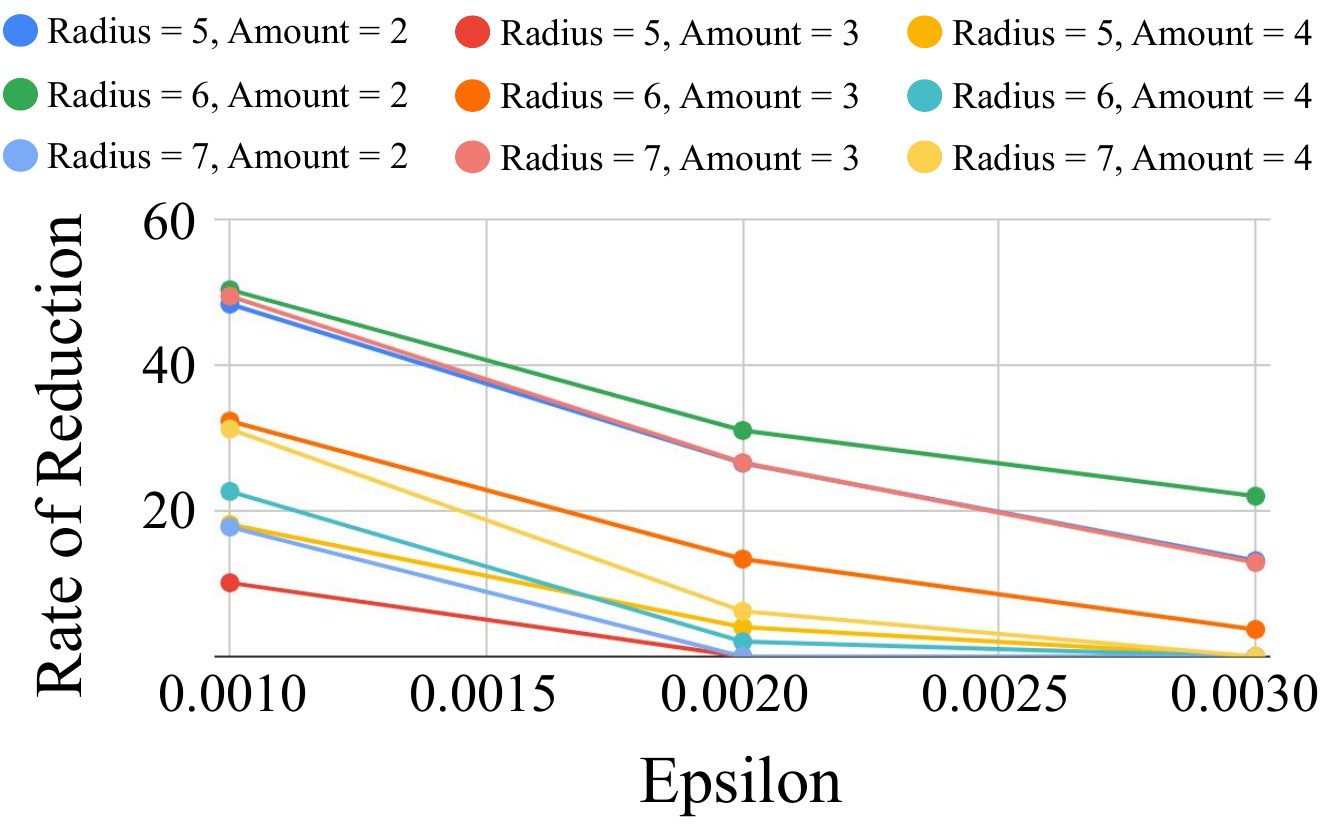}%
  }%
  \hfill
  \subcaptionbox{With segmentation}[.5\linewidth]{%
    \includegraphics[width=\linewidth]{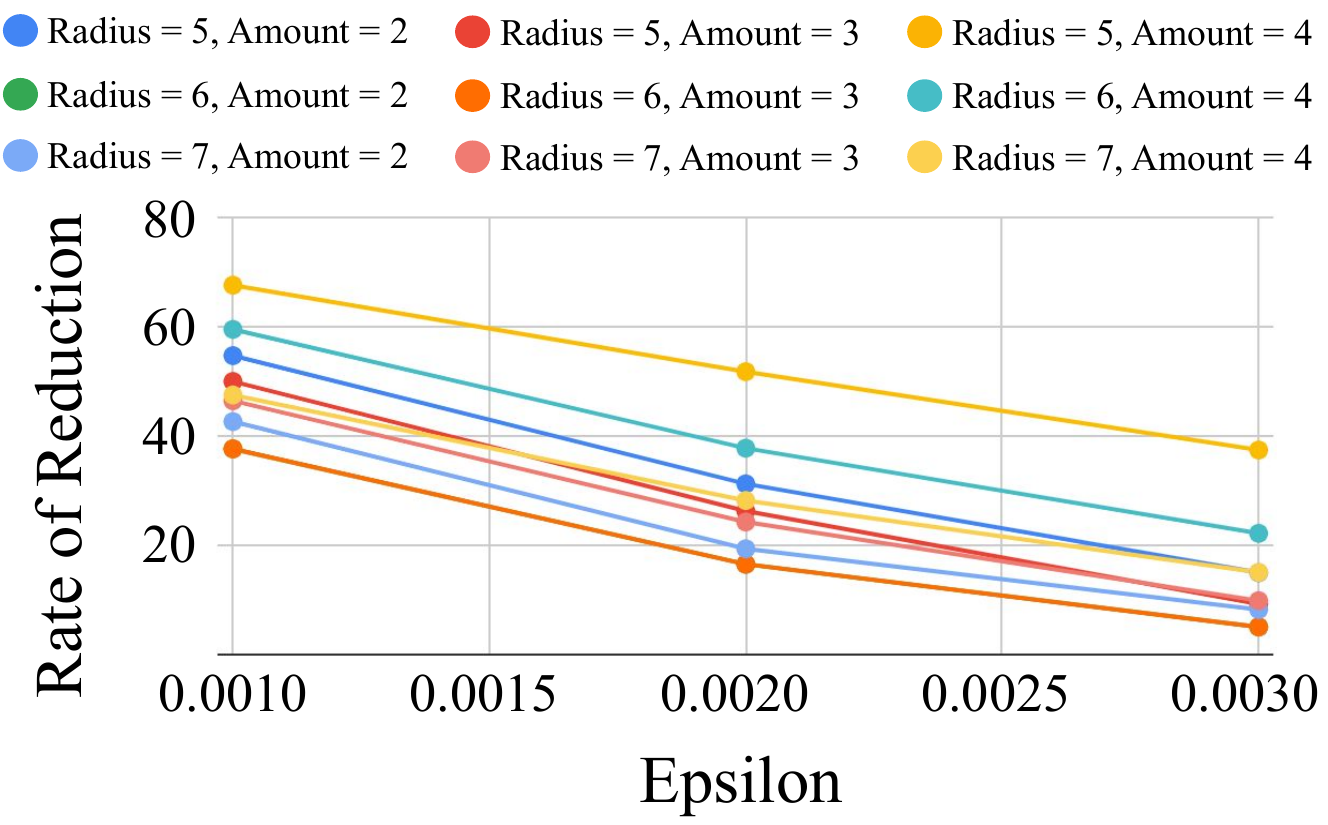}%
  }
  \caption{Rate of reduction for ViT using UM Gaussian after performing FGSM attack for epsilon values: 0.001 to 0.003.}
  \label{gaussianvit}
\end{figure}

\begin{figure}[ht]
  \subcaptionbox{Without segmentation}[.5\linewidth]{%
    \includegraphics[width=\linewidth]{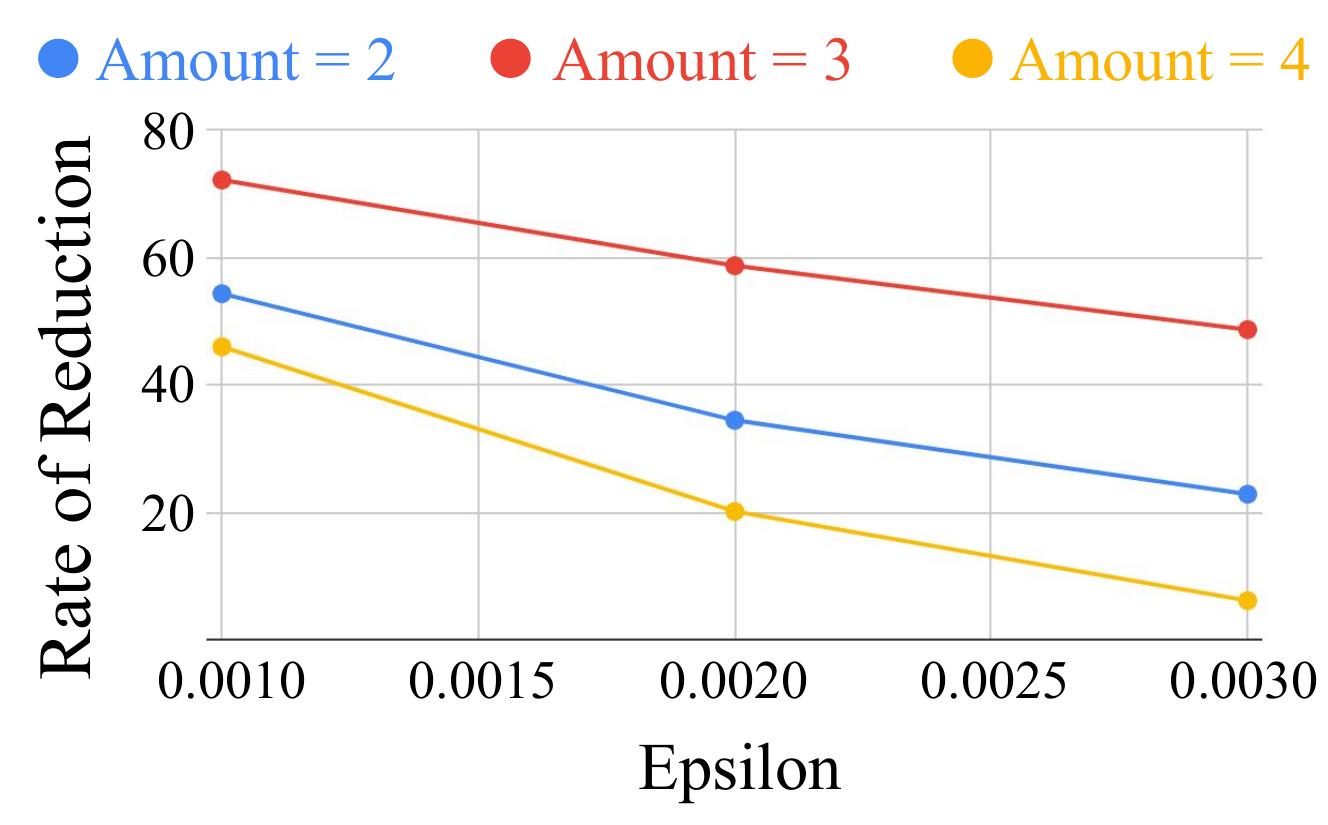}%
  }%
  \hfill
  \subcaptionbox{With segmentation}[.5\linewidth]{%
    \includegraphics[width=\linewidth]{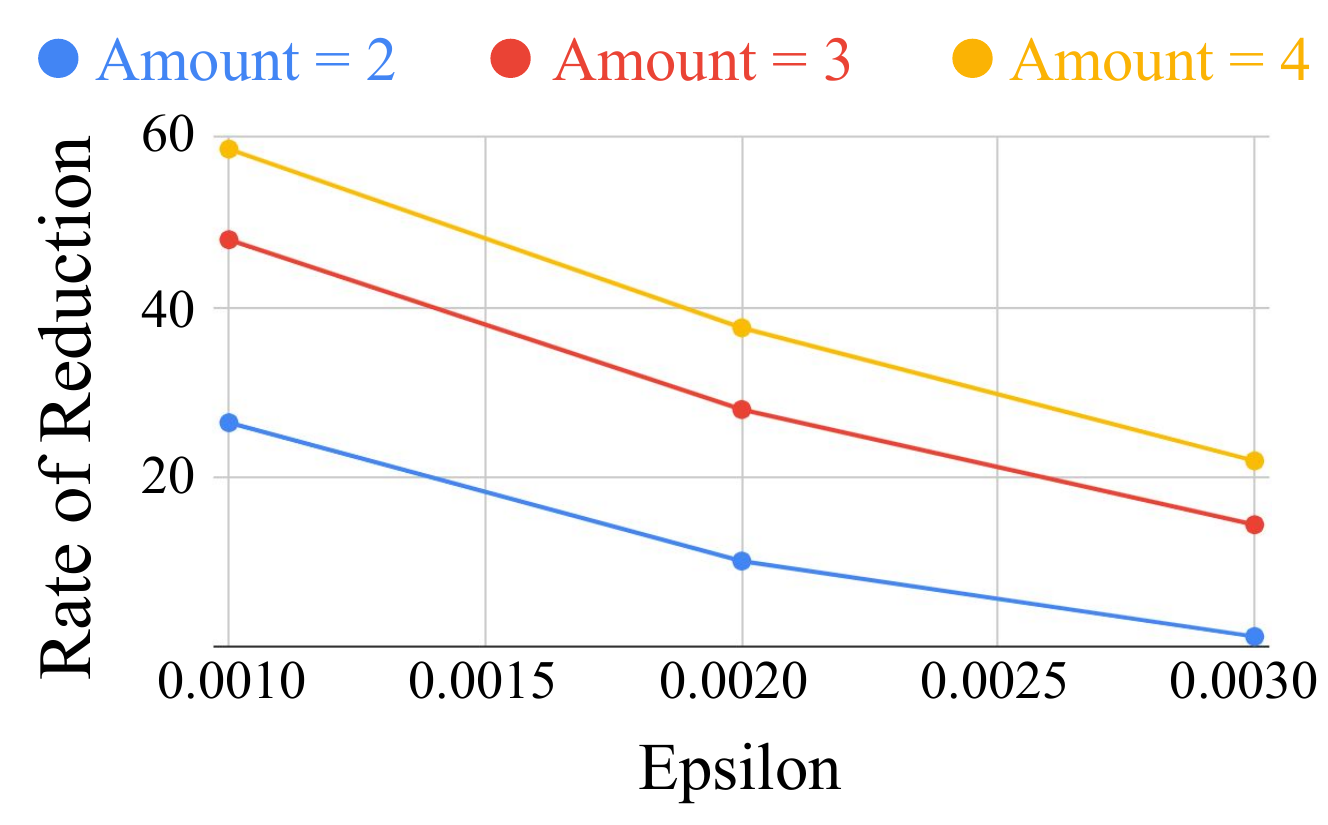}%
  }
  \caption{Rate of reduction for ViT using UM Minimum after performing FGSM attack for epsilon values: 0.001 to 0.003.}
  \label{minimumvit}
\end{figure}

\begin{figure}[ht]
  \subcaptionbox{Without segmentation}[.5\linewidth]{%
    \includegraphics[width=\linewidth]{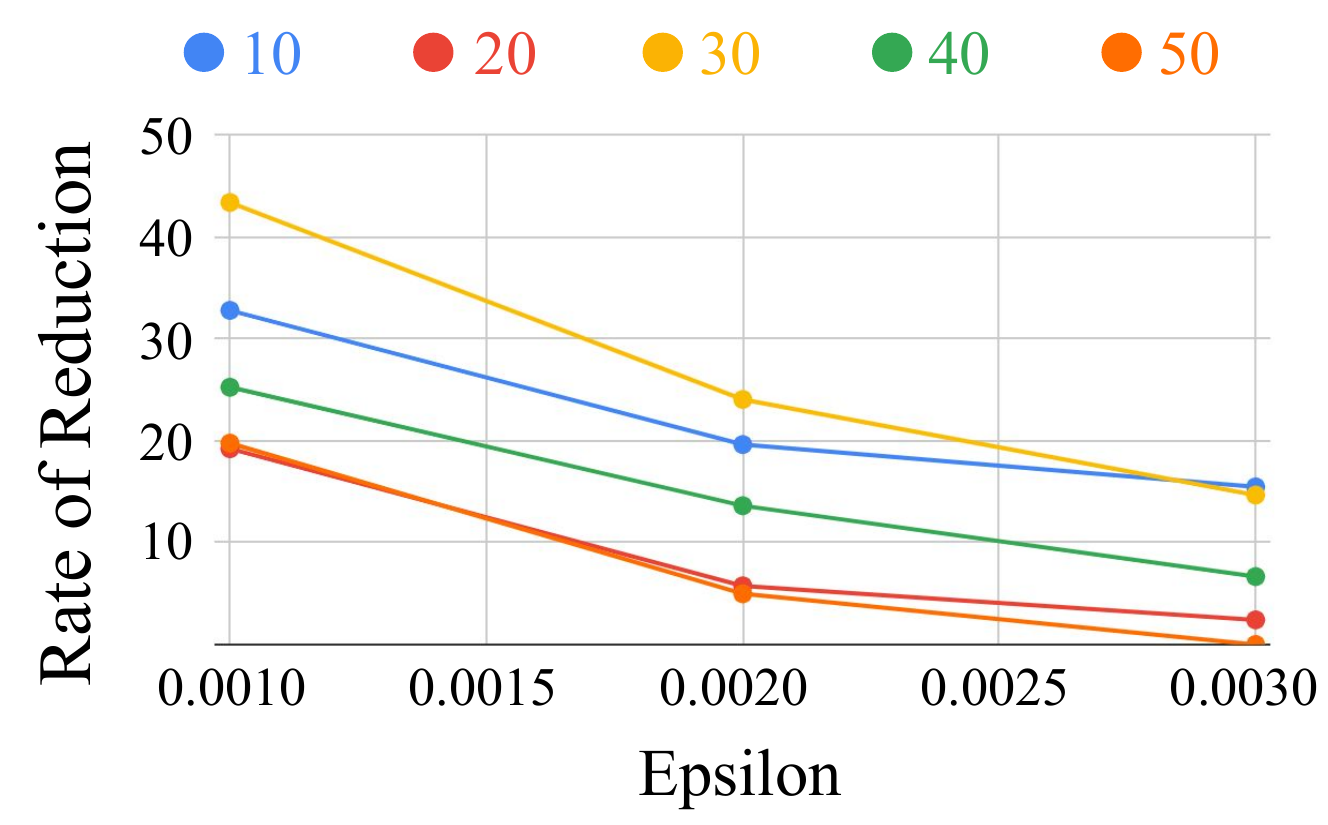}%
  }%
  \hfill
  \subcaptionbox{With segmentation}[.5\linewidth]{%
    \includegraphics[width=\linewidth]{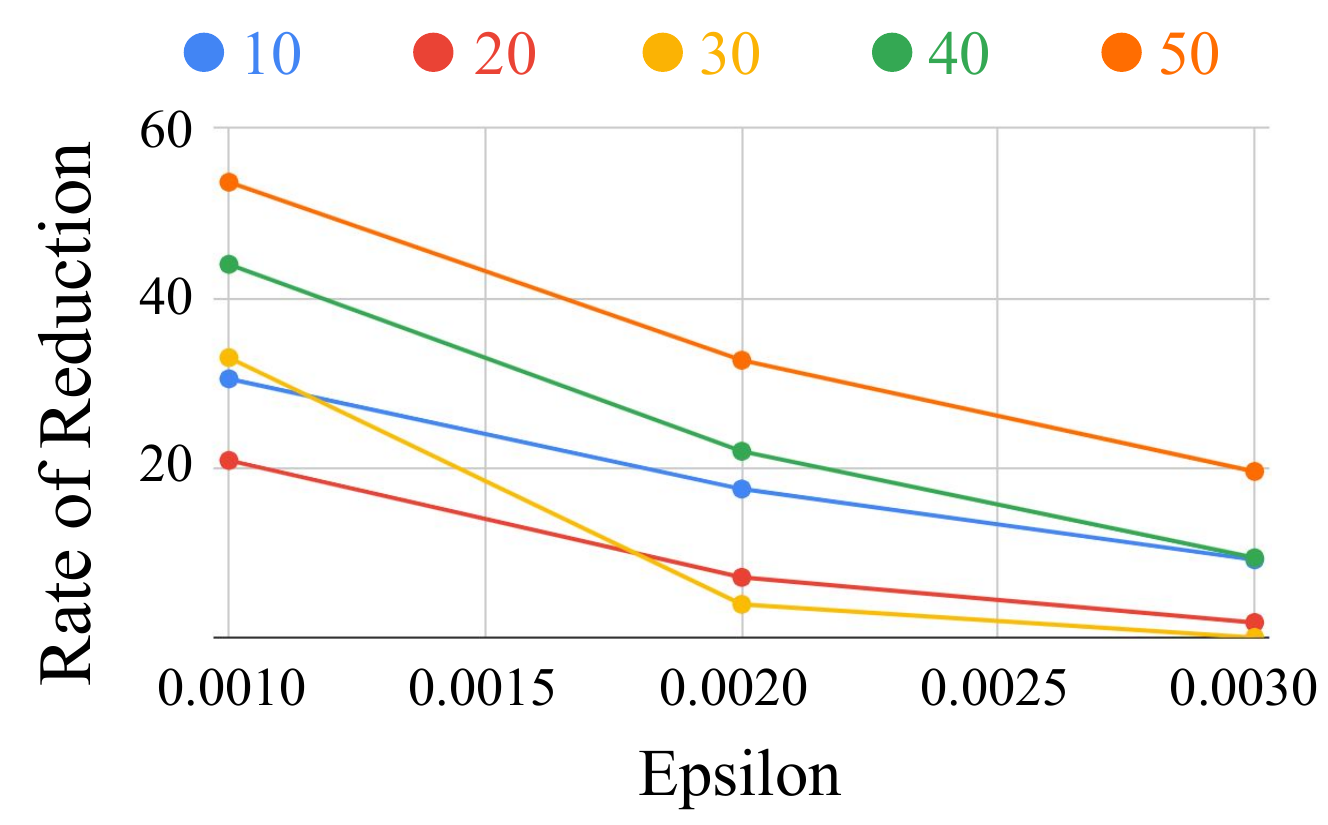}%
  }
  \caption{Rate of reduction for ViT using HFE after performing FGSM attack for epsilon values from 0.001 to 0.003.}
  \label{hfevit}
\end{figure}

\begin{figure}[ht]
  \subcaptionbox{}[.24\linewidth]{%
    \includegraphics[width=\linewidth]{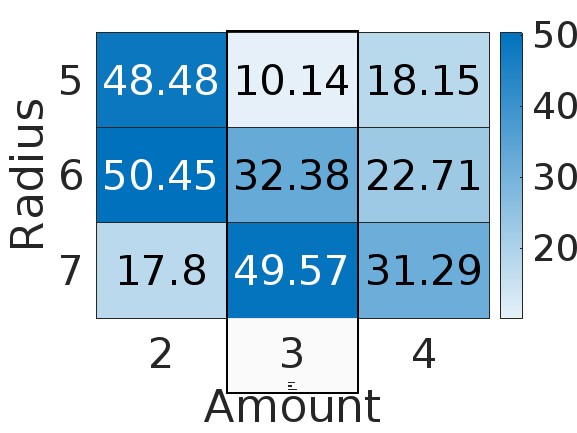}%
  }%
  \hfill
  \subcaptionbox{}[.24\linewidth]{%
    \includegraphics[width=\linewidth]{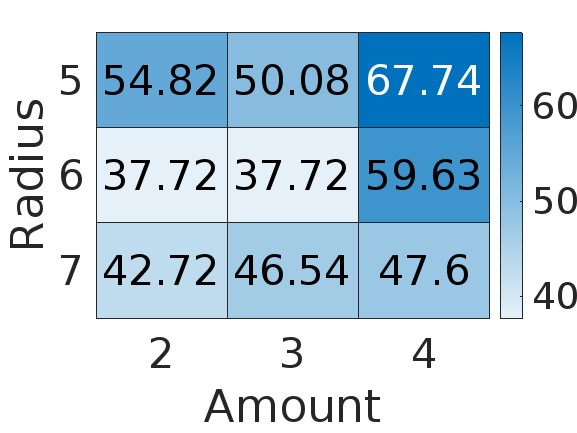}%
  }
  \subcaptionbox{}[.24\linewidth]{%
    \includegraphics[width=\linewidth]{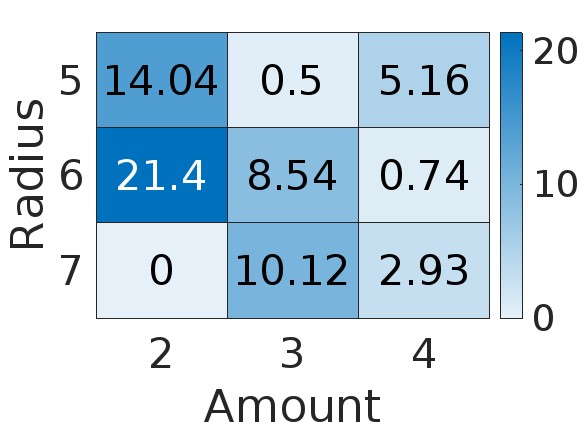}%
  }%
  \hfill
  \subcaptionbox{}[.24\linewidth]{%
    \includegraphics[width=\linewidth]{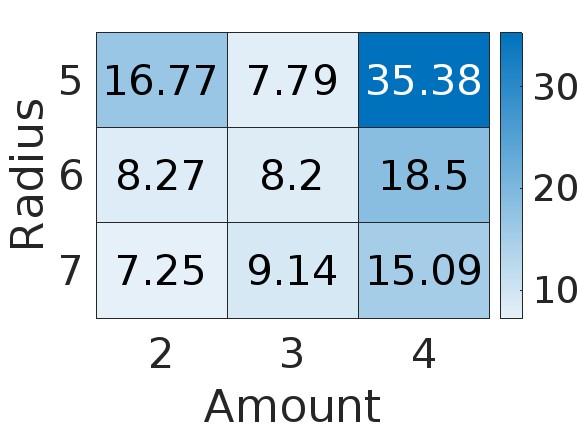}%
  }
  \caption{Rate of reduction for FGSM attack ((a)Without segmentation, (b)With segmentation) \& PGD attack ((c)Without segmentation, (d)With segmentation) for ViT using Gaussian filter for different parameter values, amount, and radius.}
  \label{heatvit}
\end{figure}

\vspace{-0.2cm}
The rate of reduction for various parameter settings for FGSM and PGD are shown in tables \ref{fgsmvit} and \ref{pgdvit} with best results highlighted in bold. 
In the case of ViT, the rate of reduction for various enhancement techniques is less compared to that of CNN. The reason is that ViTs are as such tolerant to mild perturbations in the data. These image enhancement techniques along with segmentation add to it and make the model robust and best suitable for critical applications. Figure \ref{clahevit} shows the values of the rate of reduction for various values of epsilon. Figures \ref{gaussianvit} to \ref{minimumvit} show the values for unsharp masking for different filters, gaussian, median, maximum and minimum respectively. For high-frequency emphasis filtering, Figure \ref{hfevit} shows the values for the rate of reduction without and with segmentation. Best parameter settings for Gaussian filter can be visualized in Figure \ref{heatvit}.
\vspace{-0.2cm}

\vspace{-0.2cm}

\begin{figure}[ht]
  \subcaptionbox{Without segmentation}[.5\linewidth]{%
    \includegraphics[width=\linewidth]{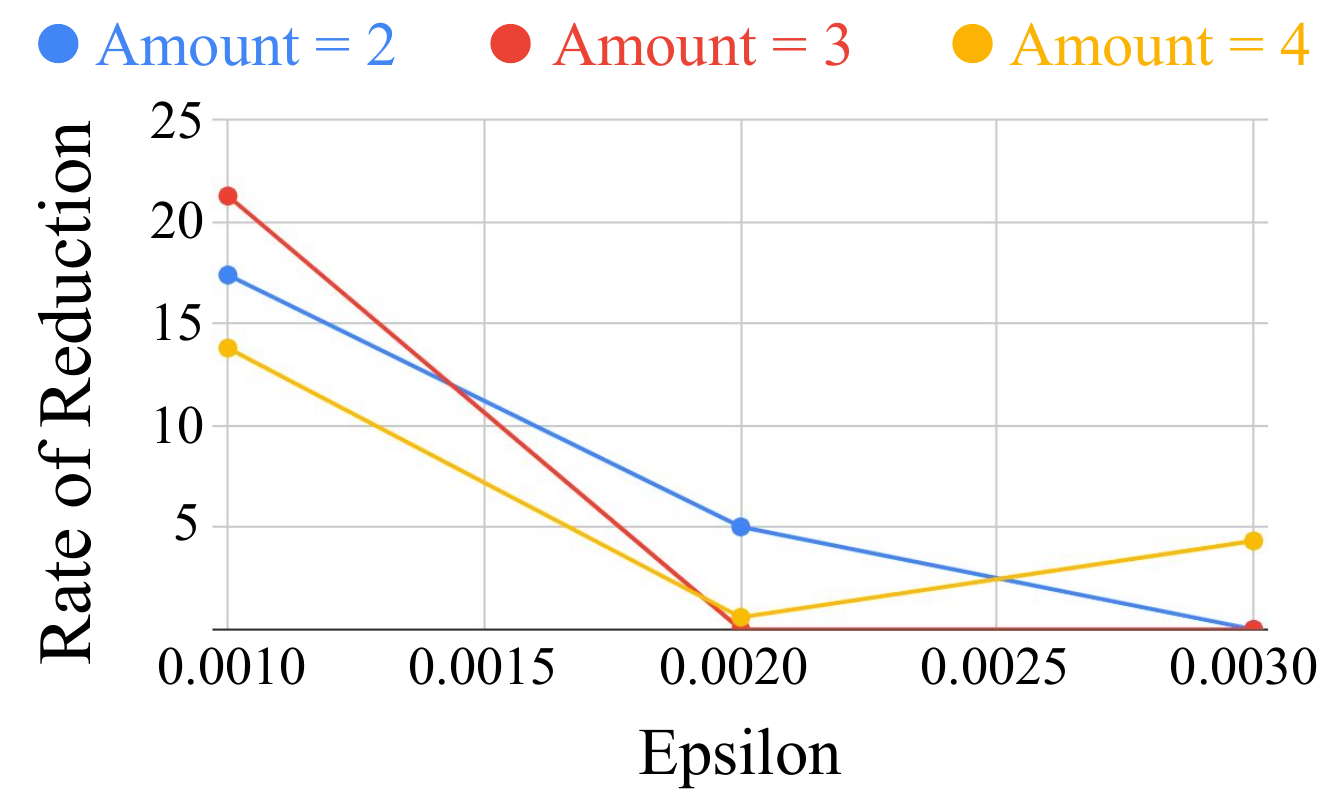}%
  }%
  \hfill
  \subcaptionbox{With segmentation}[.5\linewidth]{%
    \includegraphics[width=\linewidth]{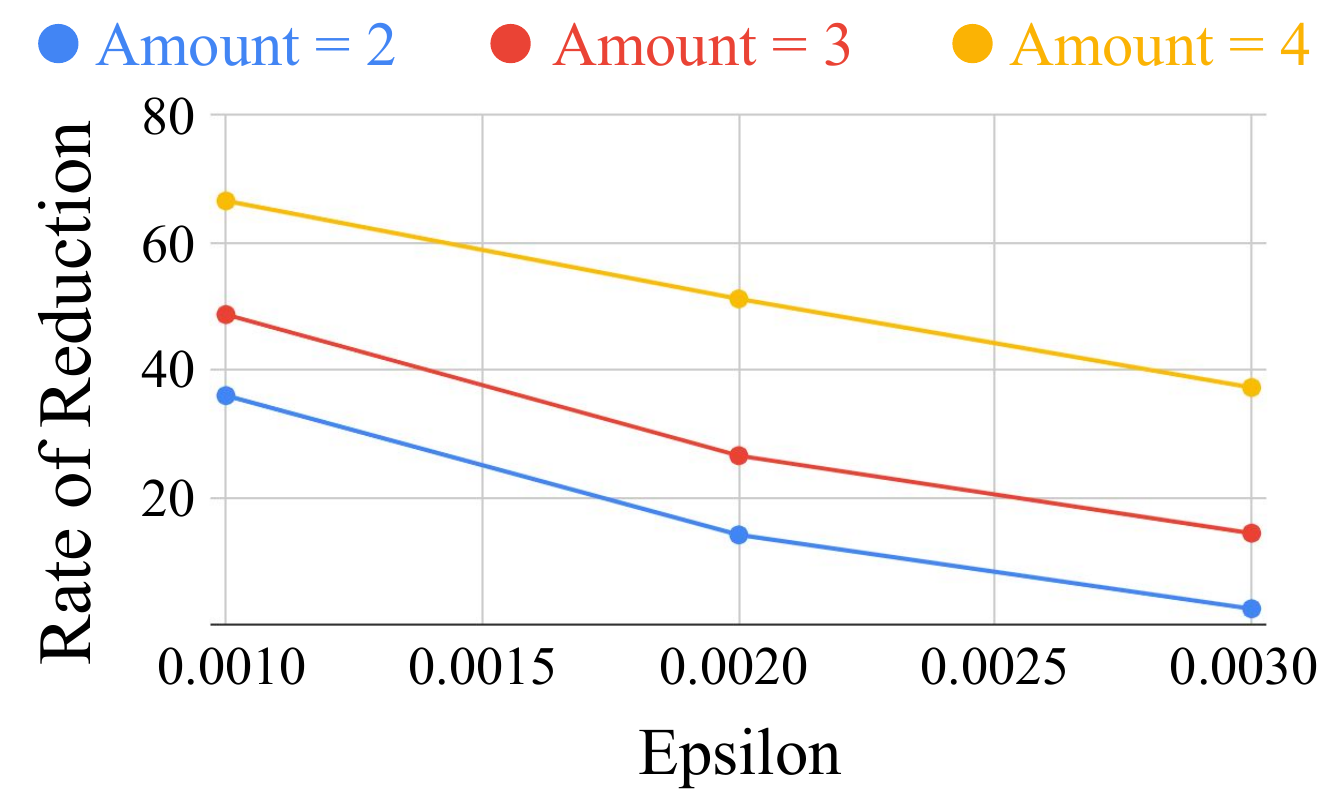}%
  }
  \caption{Rate of reduction for ViT using UM Median after performing FGSM attack for epsilon values: 0.001 to 0.003.}
\end{figure}

\begin{figure}[ht]
  \subcaptionbox{Without segmentation}[.5\linewidth]{%
    \includegraphics[width=\linewidth]{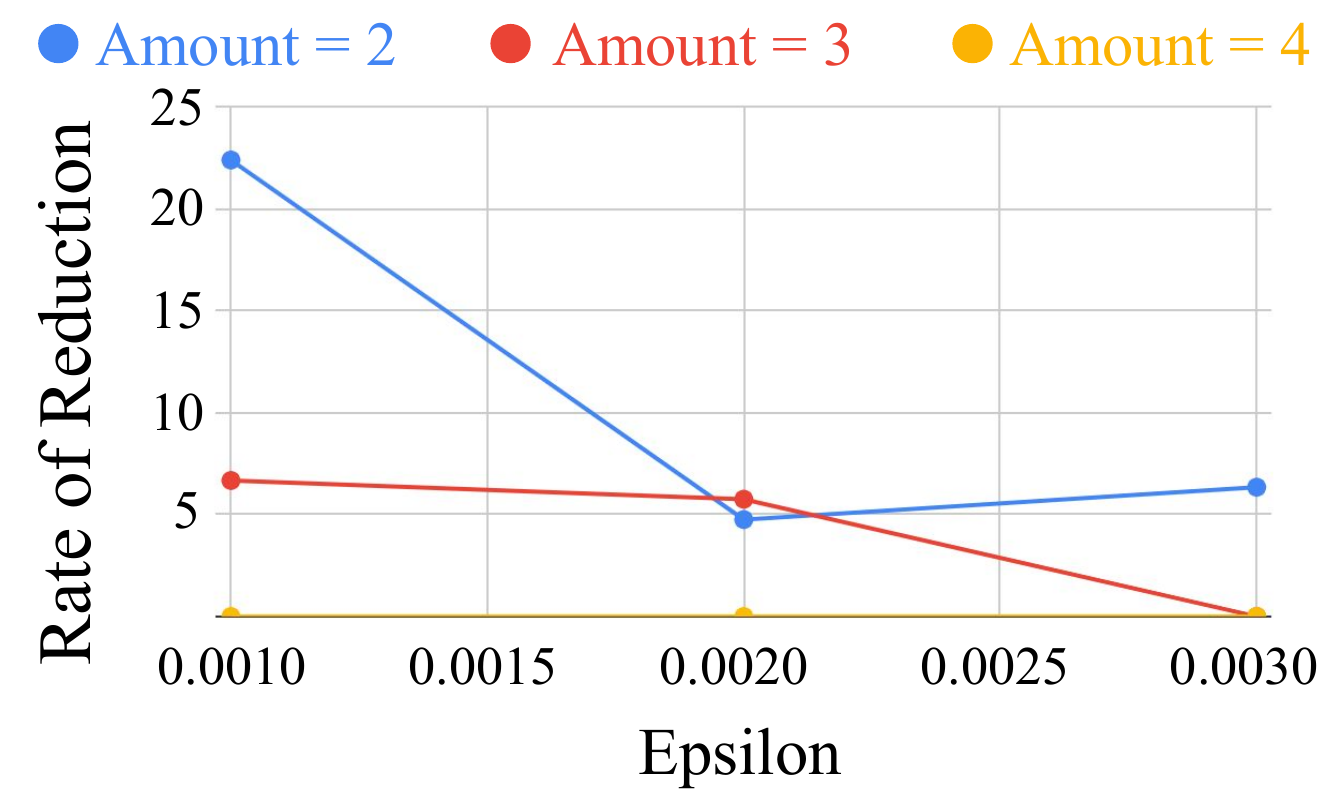}%
  }%
  \hfill
  \subcaptionbox{With segmentation}[.5\linewidth]{%
    \includegraphics[width=\linewidth]{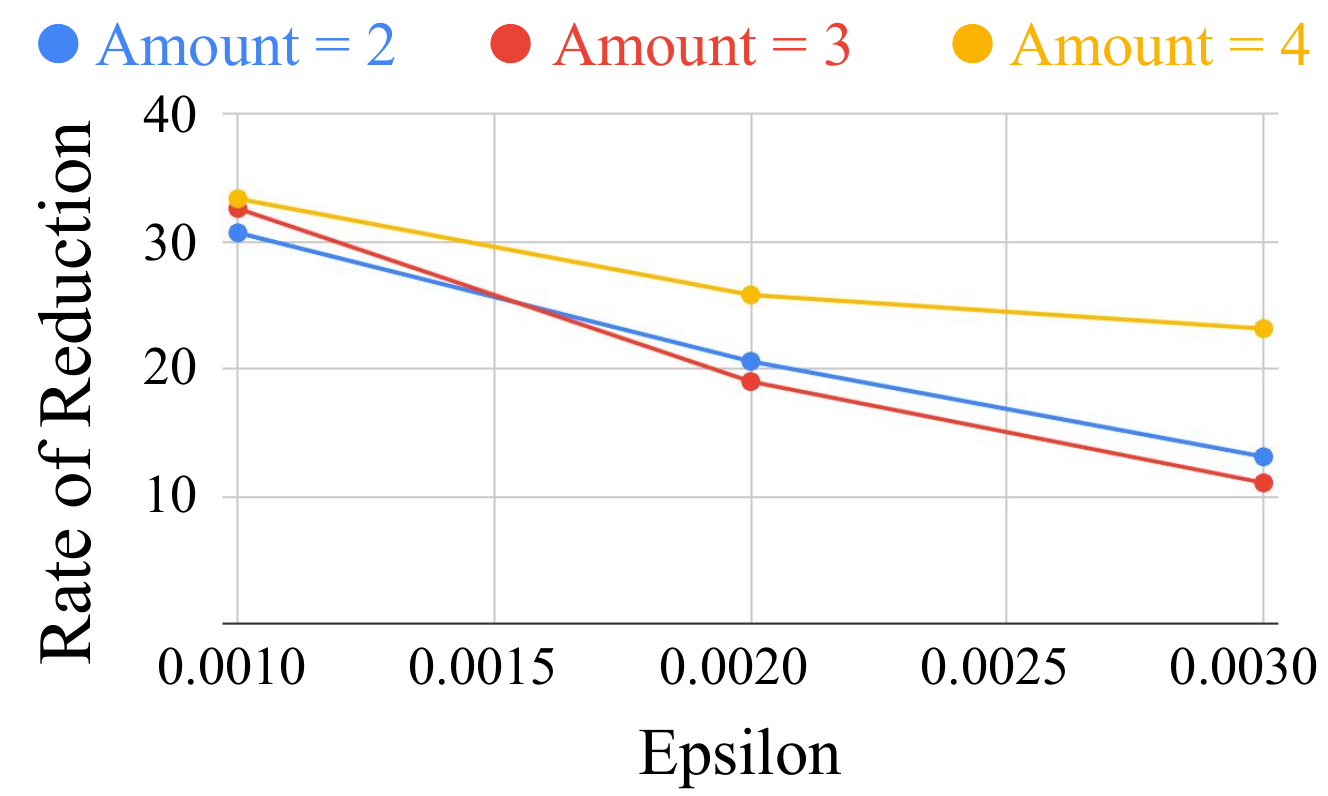}%
  }
  \caption{Rate of reduction for ViT using UM Maximum after performing FGSM attack for epsilon values: 0.001 to 0.003.}
\end{figure}


\subsection{Performance of defense mechanism in ViT-l32}
\begin{table*}[h!]
 \caption{Rate of reduction (in \%) of ViT l32 model against FGSM and PGD attack for selected parameter values.}
\renewcommand{\arraystretch}{1.3}
    \centering

    \begin{tabular}{||c|>
    {\centering}m{1cm}|>{\centering}m{1cm}|>
    {\centering}m{1cm}|>{\centering}m{1cm}|>
    {\centering}m{1cm}|>{\centering}m{1cm}|>
    {\centering}m{1cm}|>{\centering}m{1cm}|>
    {\centering}m{1cm}|>{\centering}m{1cm}|>
    {\centering}m{1cm}|>{\centering}m{1cm}|| }
    \hline
    \centering
        \multirow{2}{3em}{Attack}
        & \multicolumn{2}{c|}{CLAHE} & \multicolumn{2}{>{\centering}p{2.455cm}|}{UM-Gaussian} 
        & \multicolumn{2}{c|}{UM-Median} & \multicolumn{2}{c|}{UM-Maximum} 
        & \multicolumn{2}{c|}{UM-Minimum} & \multicolumn{2}{c||}{HFE} \tabularnewline 
        \cline{2-13}
         & Without Segmentation & With Segmentation
         & Without Segmentation & With Segmentation 
         & Without Segmentation & With Segmentation  
         & Without Segmentation & With Segmentation 
         & Without Segmentation & With Segmentation 
         & Without Segmentation & With Segmentation \tabularnewline 
      \hline \hline
        FGSM & 2.81 & 9.87 & 21.28 & 82.50 & 48.39 & 86.58 & 20.58 & 33.23 & 43.44 & 32.56 & 42.52 & 25.65  \tabularnewline 
        \hline
        PGD  & 13.82 & 11.90 & 9.70 & 79.57 & 17.30 & 79.30 & 10.56 & 18.13 & 12.35 & 23.99 & 48.26 & 80.26  \tabularnewline 
        \hline 
    \end{tabular}
    \label{vitl32table}
\end{table*}
FGSM and PGD attacks are performed in the ViT-l32 model for different image enhancement techniques and results are tabulated in Table \ref{vitl32table}. The results are shown for the best parameter set up for each enhancement technique and the best parameters which are selected are shown in Table \ref{values}.

\begin{table}[h!]
\vspace{0.2cm}
    \centering
    \caption{Best parameter values selected for experiments where r = Radius, a = Amount.}
    \begin{tabular}{||c|c|c|c||}
    \hline 
        UM-Gaussian & r = 6, a = 3 & UM-Gaussian (Seg.) & r = 5, a = 4 \\
        \hline
        UM-Median & a = 3 & UM-Median (Seg.) & a = 4 \\
        \hline
        UM-Maximum & a = 2 & UM-Maximum (Seg.) & a = 4 \\
        \hline
        UM-Minimum & a = 3 & UM-Minimum (Seg.) & a = 4 \\
        \hline
        HFE & 30 & HFE (Seg.) & 50 \\
        \hline
    \end{tabular}
    \label{values}
\end{table}

\vspace{0.2cm}
\subsection{Comparison with other preprocessing-based defenses}
\vspace{-0.1cm}

Two preprocessing-based defenses, JPEG compression and Cropping \& Resizing are studied in ViT-b32 and ViT-l32 models\cite{aldahdooh2021reveal}. Results of JPEG compression are tabulated in Table \ref{comparison} where the results from our best models are included for comparison. We compressed images with 65\% quality. Centre cropping each side of an image with 2 pixels and rescaling back to original size is done for Cropping \& Resizing technique, but the test accuracy dropped to less than 60\%.

\vspace{-0.3cm}
\subsection{Implementation in hardware}
\begin{figure}[h!]
     \centering
     \includegraphics[scale=0.55]{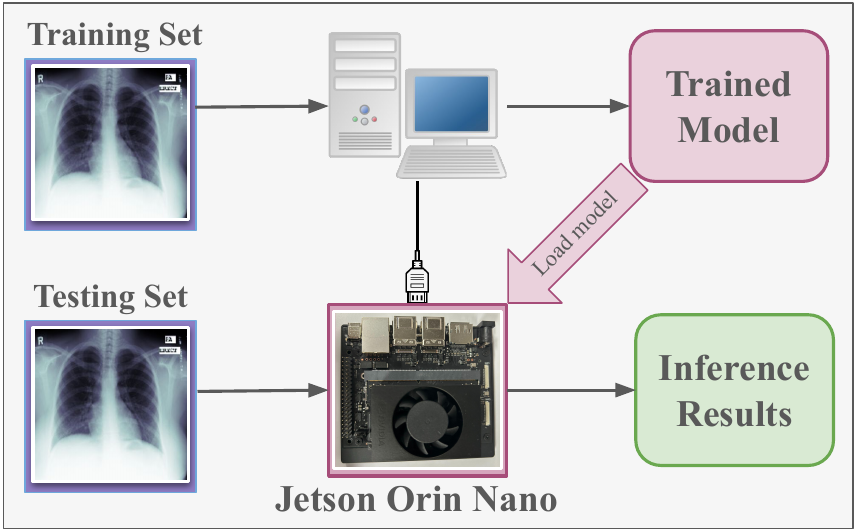}
     \caption{Inference procedure in Jetson Orin Nano board.}
     \label{board}
 \end{figure}

\vspace{-0.1cm}
For hardware implementation, we have used the Jetson Orin Nano board which has NVIDIA Ampere architecture with 1024 CUDA cores and is a computationally powerful SoC with 8 GB 128-bit LPDDR5 RAM. We have performed the inference procedure in the board as shown in Figure \ref{board} for the modified model and the results are verified. 
\vspace{0.25cm}
\begin{table}[h!]
    \centering
    \caption{Rate of reduction (in \%) for the ViT-b32 and ViT-l32 models for JPEG compression and SE pipeline.}
    \renewcommand{\arraystretch}{1.3}
    \resizebox{\columnwidth}{!}{%
    \begin{tabular}{||c|c|c|c|c||}
    \hline
       \multirow{2}{3em}{Attack}  & \multicolumn{2}{c|}{ViT-b32} & \multicolumn{2}{c||}{ViT-l32}\\
        \cline{2-5}
          & JPEG Compression & S-E Pipeline & JPEG Compression & S-E Pipeline \\
          \hline \hline
         FGSM &  9.97 & 72.22 & 17.49 & 86.58 \\
         \hline
         PGD & 13.24 & 36.25 & 2.83 & 80.26 \\
        \hline
    \end{tabular}
    }
    \label{comparison}
\end{table}
\section{Conclusion}
A ViT-based preprocessing pipeline, S-E pipeline, including segmentation and image enhancement techniques is proposed to defend against adversarial attacks. The proposed method utilizes enhancement techniques like CLAHE, UM and HFE for enhancing the image and a custom U-Net for segmentation. Experimental results using different parameter settings for image enhancement techniques show that the technique is consistently performing well against adversarial attacks. The best setup showed a rate of reduction of 72.22\% and 86.58\% for ViT-b32 and ViT-l32 models, respectively, in the effect of attacks. We also tried the same pipeline for CNN and were able to produce promising results. The entire system is tested on NVIDIA Jetson Orin Nano board for resource-constrained systems.

\vspace{-0.1cm}
\section*{Acknowledgment}



This work was supported in part by a research grant from the IHUB-NTIHAC Foundation, Indian Institute of Technology Kanpur,
India, for the project titled ‘‘Securing Deep Neural Networks against Adversarial Attacks in Medical Imaging’’ under Grant
IHUB-NTIHAC/2021/01/9; in part by the New York University Abu Dhabi (NYUAD) Center for Artificial Intelligence and Robotics
(CAIR), funded by Tamkeen through the NYUAD Research Institute under Award CG010; and in part the NYUAD Center for Cyber
Security (CCS), funded by Tamkeen through the NYUAD Research Institute under Award G1104.

\bibliography{references}

\begin{thebibliography}{10}
\providecommand{\url}[1]{#1}
\csname url@samestyle\endcsname
\providecommand{\newblock}{\relax}
\providecommand{\bibinfo}[2]{#2}
\providecommand{\BIBentrySTDinterwordspacing}{\spaceskip=0pt\relax}
\providecommand{\BIBentryALTinterwordstretchfactor}{4}
\providecommand{\BIBentryALTinterwordspacing}{\spaceskip=\fontdimen2\font plus
\BIBentryALTinterwordstretchfactor\fontdimen3\font minus \fontdimen4\font\relax}
\providecommand{\BIBforeignlanguage}[2]{{%
\expandafter\ifx\csname l@#1\endcsname\relax
\typeout{** WARNING: IEEEtran.bst: No hyphenation pattern has been}%
\typeout{** loaded for the language `#1'. Using the pattern for}%
\typeout{** the default language instead.}%
\else
\language=\csname l@#1\endcsname
\fi
#2}}
\providecommand{\BIBdecl}{\relax}
\BIBdecl

\bibitem{b3}
Q.~Li, W.~Cai, X.~Wang, Y.~Zhou, D.~D. Feng, and M.~Chen, ``Medical image classification with convolutional neural network,'' in \emph{2014 13th international conference on control automation robotics \& vision (ICARCV)}.\hskip 1em plus 0.5em minus 0.4em\relax IEEE, 2014, pp. 844--848.

\bibitem{mortazi2018automatically}
A.~Mortazi and U.~Bagci, ``Automatically designing cnn architectures for medical image segmentation,'' in \emph{Machine Learning in Medical Imaging: 9th International Workshop, MLMI 2018, Held in Conjunction with MICCAI 2018, Granada, Spain, September 16, 2018, Proceedings 9}.\hskip 1em plus 0.5em minus 0.4em\relax Springer, 2018, pp. 98--106.

\bibitem{sultana2020review}
F.~Sultana, A.~Sufian, and P.~Dutta, ``A review of object detection models based on convolutional neural network,'' \emph{Intelligent computing: image processing based applications}, pp. 1--16, 2020.

\bibitem{ma2021understanding}
X.~Ma, Y.~Niu, L.~Gu, Y.~Wang, Y.~Zhao, J.~Bailey, and F.~Lu, ``Understanding adversarial attacks on deep learning based medical image analysis systems,'' \emph{Pattern Recognition}, vol. 110, p. 107332, 2021.

\bibitem{wang2023cross}
D.~Wang, Z.~Wang, L.~Chen, H.~Xiao, and B.~Yang, ``Cross-parallel transformer: Parallel vit for medical image segmentation,'' \emph{Sensors}, vol.~23, no.~23, p. 9488, 2023.

\bibitem{li2023lvit}
Z.~Li, Y.~Li, Q.~Li, P.~Wang, D.~Guo, L.~Lu, D.~Jin, Y.~Zhang, and Q.~Hong, ``Lvit: language meets vision transformer in medical image segmentation,'' \emph{IEEE transactions on medical imaging}, 2023.

\bibitem{almalik2022self}
F.~Almalik, M.~Yaqub, and K.~Nandakumar, ``Self-ensembling vision transformer (sevit) for robust medical image classification,'' in \emph{International Conference on Medical Image Computing and Computer-Assisted Intervention}.\hskip 1em plus 0.5em minus 0.4em\relax Springer, 2022, pp. 376--386.

\bibitem{fezza2019perceptual}
S.~A. Fezza, Y.~Bakhti, W.~Hamidouche, and O.~D{\'e}forges, ``Perceptual evaluation of adversarial attacks for cnn-based image classification,'' in \emph{2019 Eleventh International Conference on Quality of Multimedia Experience (QoMEX)}.\hskip 1em plus 0.5em minus 0.4em\relax IEEE, 2019, pp. 1--6.

\bibitem{mahmood2021robustness}
K.~Mahmood, R.~Mahmood, and M.~Van~Dijk, ``On the robustness of vision transformers to adversarial examples,'' in \emph{Proceedings of the IEEE/CVF International Conference on Computer Vision}, 2021, pp. 7838--7847.

\bibitem{fgsm}
I.~J. Goodfellow, J.~Shlens, and C.~Szegedy, ``Explaining and harnessing adversarial examples,'' \emph{arXiv preprint arXiv:1412.6572}, 2014.

\bibitem{pgd}
A.~M{\k{a}}dry, A.~Makelov, L.~Schmidt, D.~Tsipras, and A.~Vladu, ``Towards deep learning models resistant to adversarial attacks,'' \emph{stat}, vol. 1050, p.~9, 2017.

\bibitem{9259112}
Y.~Lin, H.~Zhao, X.~Ma, Y.~Tu, and M.~Wang, ``Adversarial attacks in modulation recognition with convolutional neural networks,'' \emph{IEEE Transactions on Reliability}, vol.~70, no.~1, pp. 389--401, 2021.

\bibitem{shao2022adversarial}
R.~Shao, Z.~Shi, J.~Yi, P.-Y. Chen, and C.-J. Hsieh, ``On the adversarial robustness of vision transformers,'' 2022.

\bibitem{9420266}
H.~Qiu, Y.~Zeng, Q.~Zheng, S.~Guo, T.~Zhang, and H.~Li, ``An efficient preprocessing-based approach to mitigate advanced adversarial attacks,'' \emph{IEEE Transactions on Computers}, pp. 1--1, 2021.

\bibitem{chang2023enhancing}
Y.~Chang, H.~Zhao, and W.~Wang, ``Enhancing the robustness of vision transformer defense against adversarial attacks based on squeeze-and-excitation module,'' \emph{PeerJ Computer Science}, vol.~9, p. e1197, 2023.

\bibitem{finlayson2018adversarial}
S.~G. Finlayson, H.~W. Chung, I.~S. Kohane, and A.~L. Beam, ``Adversarial attacks against medical deep learning systems,'' \emph{arXiv preprint arXiv:1804.05296}, 2018.

\bibitem{paschali2018generalizability}
M.~Paschali, S.~Conjeti, F.~Navarro, and N.~Navab, ``Generalizability vs. robustness: investigating medical imaging networks using adversarial examples,'' in \emph{Medical Image Computing and Computer Assisted Intervention--MICCAI 2018: 21st International Conference, Granada, Spain, September 16-20, 2018, Proceedings, Part I}.\hskip 1em plus 0.5em minus 0.4em\relax Springer, 2018, pp. 493--501.

\bibitem{kaviani2022adversarial}
S.~Kaviani, K.~J. Han, and I.~Sohn, ``Adversarial attacks and defenses on ai in medical imaging informatics: A survey,'' \emph{Expert Systems with Applications}, vol. 198, p. 116815, 2022.

\bibitem{dziugaite2016study}
G.~K. Dziugaite, Z.~Ghahramani, and D.~M. Roy, ``A study of the effect of jpg compression on adversarial images,'' \emph{arXiv preprint arXiv:1608.00853}, 2016.

\bibitem{graese2016assessing}
A.~Graese, A.~Rozsa, and T.~E. Boult, ``Assessing threat of adversarial examples on deep neural networks,'' in \emph{2016 15th IEEE International Conference on Machine Learning and Applications (ICMLA)}.\hskip 1em plus 0.5em minus 0.4em\relax IEEE, 2016, pp. 69--74.

\bibitem{Xu_2018}
\BIBentryALTinterwordspacing
W.~Xu, D.~Evans, and Y.~Qi, ``Feature squeezing: Detecting adversarial examples in deep neural networks,'' in \emph{Proceedings 2018 Network and Distributed System Security Symposium}, ser. NDSS 2018.\hskip 1em plus 0.5em minus 0.4em\relax Internet Society, 2018. [Online]. Available: \url{http://dx.doi.org/10.14722/ndss.2018.23198}
\BIBentrySTDinterwordspacing

\bibitem{guo2017countering}
C.~Guo, M.~Rana, M.~Cisse, and L.~Van Der~Maaten, ``Countering adversarial images using input transformations,'' \emph{arXiv preprint arXiv:1711.00117}, 2017.

\bibitem{aldahdooh2021reveal}
A.~Aldahdooh, W.~Hamidouche, and O.~Deforges, ``Reveal of vision transformers robustness against adversarial attacks,'' \emph{arXiv preprint arXiv:2106.03734}, 2021.

\bibitem{mo2022adversarial}
Y.~Mo, D.~Wu, Y.~Wang, Y.~Guo, and Y.~Wang, ``When adversarial training meets vision transformers: Recipes from training to architecture,'' \emph{Advances in Neural Information Processing Systems}, vol.~35, pp. 18\,599--18\,611, 2022.

\bibitem{electronics11203370}
\BIBentryALTinterwordspacing
Y.~Chang, H.~Zhao, and W.~Wang, ``Ask-vit: A model with improved vit robustness through incorporating sk modules using adversarial training,'' \emph{Electronics}, vol.~11, no.~20, 2022. [Online]. Available: \url{https://www.mdpi.com/2079-9292/11/20/3370}
\BIBentrySTDinterwordspacing

\bibitem{imam2023enhancing}
R.~Imam, M.~Huzaifa, and M.~E.-A. Azz, ``On enhancing the robustness of vision transformers: Defensive diffusion,'' 2023.

\bibitem{MANZARI2023106791}
\BIBentryALTinterwordspacing
O.~N. Manzari, H.~Ahmadabadi, H.~Kashiani, S.~B. Shokouhi, and A.~Ayatollahi, ``Medvit: A robust vision transformer for generalized medical image classification,'' \emph{Computers in Biology and Medicine}, vol. 157, p. 106791, 2023. [Online]. Available: \url{https://www.sciencedirect.com/science/article/pii/S0010482523002561}
\BIBentrySTDinterwordspacing

\bibitem{jaeger2014two}
S.~Jaeger, S.~Candemir, S.~Antani, Y.-X.~J. W{\'a}ng, P.-X. Lu, and G.~Thoma, ``Two public chest x-ray datasets for computer-aided screening of pulmonary diseases,'' \emph{Quantitative imaging in medicine and surgery}, vol.~4, no.~6, p. 475, 2014.

\end{thebibliography}
\bibliographystyle{IEEEtran}



\end{document}